\pdfoutput=1
\documentclass[10pt,twocolumn,letterpaper]{article}

\usepackage{iccv}
\usepackage{times}
\usepackage{epsfig}
\usepackage{graphicx}
\usepackage{amsmath}
\usepackage{amssymb}

\usepackage[pagebackref=true,breaklinks=true,letterpaper=true,colorlinks,bookmarks=false]{hyperref}

\iccvfinalcopy 

\def\iccvPaperID{9013} 

\ificcvfinal\fi

\newcommand{\loss}{\mathcal{L}}
\DeclareMathOperator*{\argmax}{arg\,max}
\DeclareMathOperator*{\argmin}{arg\,min}
\usepackage{bm}
\usepackage{multicol}
\usepackage{multirow}
\usepackage{booktabs}
\usepackage{xspace}
\usepackage[title]{appendix}
\usepackage[ruled,vlined]{algorithm2e}
\let\oldnl\nl
\newcommand{\nonl}{\renewcommand{\nl}{\let\nl\oldnl}}
\SetKwInOut{Parameter}{Parameter}

\begin{document}

\title{Discovering Robust Convolutional Architecture at Targeted Capacity: \\A Multi-Shot Approach}

\author{Xuefei Ning$^1$\thanks{Equal contribution.} \and Junbo Zhao$^{1*}$ \and Wenshuo Li$^1$ \and Tianchen Zhao$^{1}$ \and Yin Zheng$^2$ \and Huazhong Yang$^1$\thanks{Corresponding authors.} \and Yu Wang$^{1\dagger}$}

\twocolumn[
\begin{@twocolumnfalse}
\maketitle
\begin{center}
{\normalsize 
\vspace{-25pt}
$^1$Department of Electronic Engineering, Tsinghua University\\
$^2$Wechat group, Tencent\\}
\end{center}
\end{@twocolumnfalse}
]
{
  \renewcommand{\thefootnote}%
    {\fnsymbol{footnote}}
    \footnotetext[1]{Equal contribution.}
      \renewcommand{\thefootnote}%
    {\fnsymbol{footnote}}
    \footnotetext[2]{Corresponding authors. }
    }


\ificcvfinal\thispagestyle{empty}\fi

\begin{abstract}
  Convolutional neural networks (CNNs) are vulnerable to adversarial examples, and studies show that increasing the model capacity of an architecture topology (e.g., width expansion) can bring consistent robustness improvements. 
  This reveals a clear robustness-efficiency trade-off that should be considered in architecture design.
  In this paper, considering scenarios with capacity budget, we aim to discover adversarially robust architecture at targeted capacities.
  Recent studies employed one-shot neural architecture search (NAS) to discover robust architectures.
  However, since the capacities of different topologies cannot be aligned in the search process, one-shot NAS methods favor topologies with larger capacities in the supernet. And the discovered topology might be suboptimal when augmented to the targeted capacity.
  We propose a novel multi-shot NAS method to address this issue and explicitly search for robust architectures at targeted capacities. 
   At the targeted FLOPs of 2000M, the discovered MSRobNet-2000 outperforms the recent NAS-discovered architecture RobNet-large under various criteria by a large margin of 4\%-7\%. And at the targeted FLOPs of 1560M, MSRobNet-1560 surpasses another NAS-discovered architecture RobNet-free by 2.3\% and 1.3\% in the clean and PGD$^7$ accuracies, respectively. All codes are available at \url{
https://github.com/walkerning/aw\_nas}.
\end{abstract}

\section{Introduction}

\noindent Convolutional neural networks (CNNs) are known to be vulnerable to adversarial examples (i.e., adversarially crafted imperceptible perturbations)~\cite{szegedy2013intriguing}. For defending against adversarial examples, extensive efforts are devoted to designing training or regularization techniques~\cite{cisse2017parseval,madry2018towards,zhang2019theoretically} and introducing special modules (e.g., randomness injection~\cite{guo2018countering}, generative model~\cite{schott2019towards,song2018pixeldefend}). Currently, only a few studies~\cite{guo2019meets,Chen2020AntiBanditNA,dong2020adversarially} have explored the robustness characteristics from the architectural aspect.  

Previous studies~\cite{madry2018towards} have observed that increasing the capacity of topology by width expansion brings consistent robustness improvements. Therefore, one can simply expand the architecture width for higher robustness at the cost of increasing model capacity.
Nevertheless, in actual deployment, there usually exists a capacity budget requirement on the architecture. 
To discover superior architectures when augmented to the capacity budget by width expansion, for the first time, 
we propose to employ neural architecture search (NAS) to search for adversarially robust architectures \textbf{at targeted capacities}.




Currently, parameter-sharing techniques are widely used in NAS methods to boost the search efficiency~\cite{pham2018efficient,darts}.
In parameter-sharing NAS, candidate architectures are evaluated directly using the weights in an over-parametrized super network (i.e., supernet).
After a topology is found,
model augmentation along the width or depth dimension is usually applied to construct a larger final architecture. One-shot NAS is a specific type of parameter-sharing NAS, in which the search process is decoupled into two phases: 1) Train a supernet; 2) Search while using the supernet weights to evaluate candidate architectures. 

\begin{figure}[bt]
  \centering
  \includegraphics[width=0.46\textwidth]{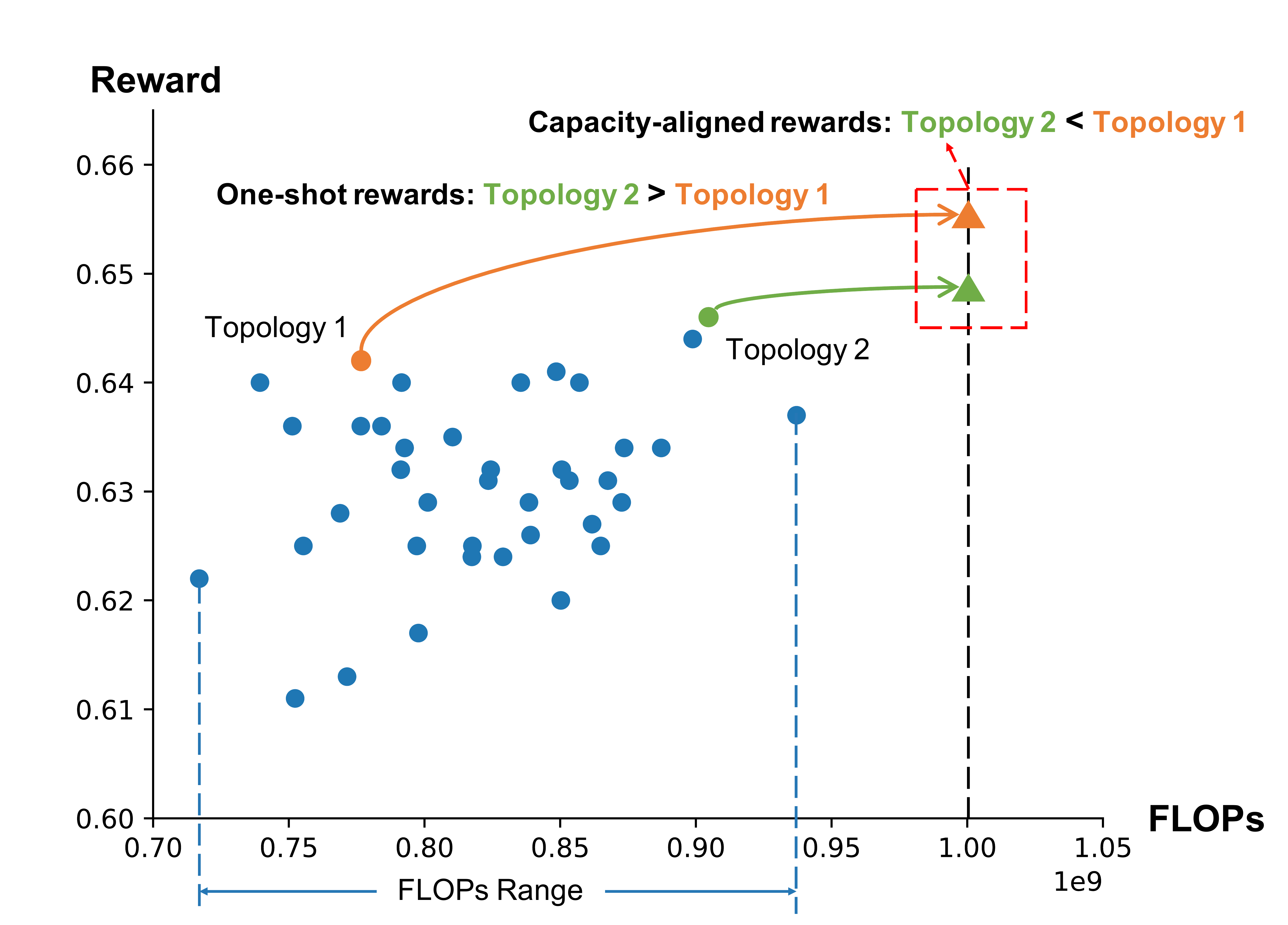}
  \caption{A motivating illustration. The FLOPs range of topologies in supernets is large. When evaluated in the supernet, topology 2 (green) is better than topology 1 (orange). However, when aligned to the same capacity, topology 1 is  better than topology 2.}
  \label{fig:motivating} 
\end{figure}

Some studies~\cite{Chen2020AntiBanditNA,dong2020adversarially} that apply parameter-sharing NAS for robustness ignore the robustness-efficiency trade-off. Guo et. al.~\cite{guo2019meets} 
consider the capacity issue in their one-shot NAS flow, but the search is not targeting certain capacities. And there exists a problem in one-shot NAS that hinders its effectiveness in targeting certain capacities: the model capacity of different topologies cannot be aligned in the supernet, and one-shot NAS favors larger topologies which can be suboptimal at the targeted capacities.
As illustrated in Fig.~\ref{fig:motivating}, Topology 2 is discovered by one-shot NAS since it has a high one-shot reward. However, Topology 1 with a lower reward and capacity in the supernet might outperform Topology 2 when they are aligned to the same capacity.




In this paper, we propose a novel multi-shot NAS method to \textbf{search for adversarially robust architectures at targeted capacities}, while taking full advantage of the parameter sharing technique.
The core of the multi-shot NAS method is to inter- or extra-polate the performances evaluated by multiple ``one-shot'' supernets of different sizes to estimate the ``multi-shot'' reward at the targeted capacity.
The contributions of this paper are as follows:
\begin{itemize}
    \item We propose a novel multi-shot NAS method to search for adversarially robust architectures at targeted capacity. We verify that the multi-shot evaluation strategy can bridge the correlation gap brought by the capacity misalignment issue in one-shot supernets.
    \item To improve the efficiency of the NAS for robustness process, we use the FGSM reward as a reliable and efficient proxy of more accurate robustness criteria, and conduct a ranking correlation analysis of various robustness criteria to verify its rationality.
    \item Experimental results show that at targeted capacities, our discovered MSRobNet architectures outperform manually-designed ones and recent NAS-discovered ones significantly. 
\end{itemize}



\section{Related Work}\label{sec:rw}

\subsection{Adversarial Attacks and Defenses}


\noindent\textbf{Adversarial Attacks } Current studies usually run adversarial attacks to evaluate the robustness of models.
Commonly used white-box attacking methods can be classified into local approximation based ones and optimization based ones. Local approximation based methods craft adversarial inputs by following update rules derived with certain local approximations, including the fast gradient sign method (FGSM) and its variants~\cite{Goodfellow2015ExplainingAH,kurakin2016adversarial}, saliency-based~\cite{papernot2016limitations} methods, local decision boundary method~\cite{moosavi2016deepfool}. On the other hand, optimization-based attacking methods formulate the attack as one or multiple optimization problems and solve the optimization problems approximately, of which the most popular ones are the C\&W attack and its variants~\cite{carlini2017towards,carlini2017adversarial}.

\noindent\textbf{Adversarial Defenses } Existing adversarial defenses can be classified into several categories: 1) adversarial example detection~\cite{metzen2017detecting,carlini2017adversarial}; 2) input mapping or processing~\cite{song2018pixeldefend,guo2018countering}; 3) regularization techniques~\cite{papernot2016distillation,cisse2017parseval}; 4) adversarial training~\cite{madry2018towards,zhang2019theoretically}.
Among the extensive literature, many defenses are proved to be not useful to stronger attacks~\cite{carlini2017adversarial,athalye2018obfuscated}, and adversarial training is acknowledged as the most effective defense technique.

\noindent\textbf{Robust CNN Architecture Design } In contrast to the aforementioned defenses, we aim to improve the adversarial robustness of CNNs 
by designing more robust neural architectures. The two most related studies to our work are 1) Guo et al.~\cite{guo2019meets} employ one-shot NAS to investigate the architecture patterns that are beneficial to adversarial robustness, and find that densely connected pattern is beneficial.
They analyze architectures in three coarsely-partitioned capacity range (i.e., small, medium, large), by randomly sampling 100 architecture for each range. In contrast, we develop the multi-shot evaluation strategy to estimate the reward at targeted capacities, and conduct a guided search using the estimated reward.
2) Chen et al.~\cite{Chen2020AntiBanditNA} employ an anti-bandit algorithm to improve the search efficiency. To accelerate the search process, they employ FGSM adversarial training in the search process, while the final model robustness is evaluated with PGD attacks. However, we find that using FGSM adversarial training during the search process will result in uncorrelated evaluation (See Sec.~\ref{sec:method_fgsm}).

\subsection{Neural Architecture Search}
\noindent\textbf{Problem Definition } Neural architecture search has been recently employed to automatically discover neural architectures for various tasks~\cite{zoph2016neural,pham2018efficient,darts,real2019regularized}. The basic formalization of the NAS problem is in Eq.~\ref{eq:nas_formalization}, and many studies focus on approximately solving this problem efficiently.

\begin{equation}
    \begin{aligned}
&\mbox{max}_{\alpha \in \mathcal{A}}\quad E_{x_v \sim D_v} [R(x_v, \mbox{Net}(\alpha,w^*(\alpha)))]\\
&\mbox{s.t. } w^*(\alpha) = \mbox{argmin}_{w} E_{x_t \sim D_t} [\loss(x_t, \mbox{Net}(\alpha,w))],
\end{aligned}
\label{eq:nas_formalization}
\end{equation}
where $\mathcal{A}$ denotes the architecture search space; $x_v \sim D_t, x_t \sim D_v$ denote the data sampled from the training and validation dataset, respectively; $R$ denotes the reward used to instruct the search process, and $\loss$ denotes the loss function for backpropagation to train the weights $w$.

The vanilla NAS algorithm~\cite{zoph2016neural} is extremely slow, since thousands of architectures need to be evaluated, and the evaluation of each architecture involves a separate training process to get $w^*(\alpha)$.

\noindent\textbf{Parameter-sharing Techniques and One-shot NAS } 
A widely-used technique to accelerate the evaluation of each individual architecture is to use parameter-sharing techniques~\cite{pham2018efficient,darts}. They construct a super network such that all architectures in the search space are sub-architectures of the supernet. Then all architectures can be evaluated using a subset of the weights in the supernet. In this way, the training costs of architectures are amortized to only train an over-parametrized supernet.
One-shot NAS~\cite{Bender2018UnderstandingAS} is a specific type of parameter-sharing based NAS method, in which the supernet training process and the architecture search process are decoupled. 

\noindent\textbf{Predictor-based Controller } To reduce the number of architectures needed to be evaluated (i.e., improve the sample efficiency of the search process), predictor-based controllers~\cite{nao2018,kandasamy2018bayesian,gates} predict the architectures' performances, and only sample promising architectures. In this way, fewer architectures need to go through the relatively expensive evaluation procedure, which includes a separate inference process on the validation data split and sometimes a separate training process on the training data split.

\section{Preliminary: One-shot NAS Workflow}\label{sec:motivation}
A typical one-shot NAS workflow goes as follows: 1) Train a supernet; 2) Conduct architecture search while using the supernet weights to evaluate candidate architectures. 3) In cell-based search spaces~\cite{pham2018efficient,darts}, the model augmentation technique (increasing the channel/layer number) is usually applied on the discovered topology to construct a large final architecture. 4) Finally, the augmented architecture is trained on the dataset.

Previous work~\cite{madry2018towards} has shown that expanding the model width brings consistent robustness improvements at the cost of increasing capacity.
Then, a natural question is that can we search for a topology that is superior at certain capacities with one-shot NAS?
Unfortunately, since the parameters of different architectures are shared, their capacity cannot be easily aligned. Specifically, the FLOPs (number of floating-point operations) of different topologies can vary in a large range in the supernet
(e.g., for our search space, 304M-1344M  with 44 init channels).
Topologies with larger capacity in the supernet tend to have better one-shot performances. However, these topologies might no longer outperform those with smaller capacity when augmented to a targeted capacity. A motivating illustration is shown in Fig.~\ref{fig:motivating}.

\begin{figure*}[tb]
  \centering
  \includegraphics[width=0.99\textwidth]{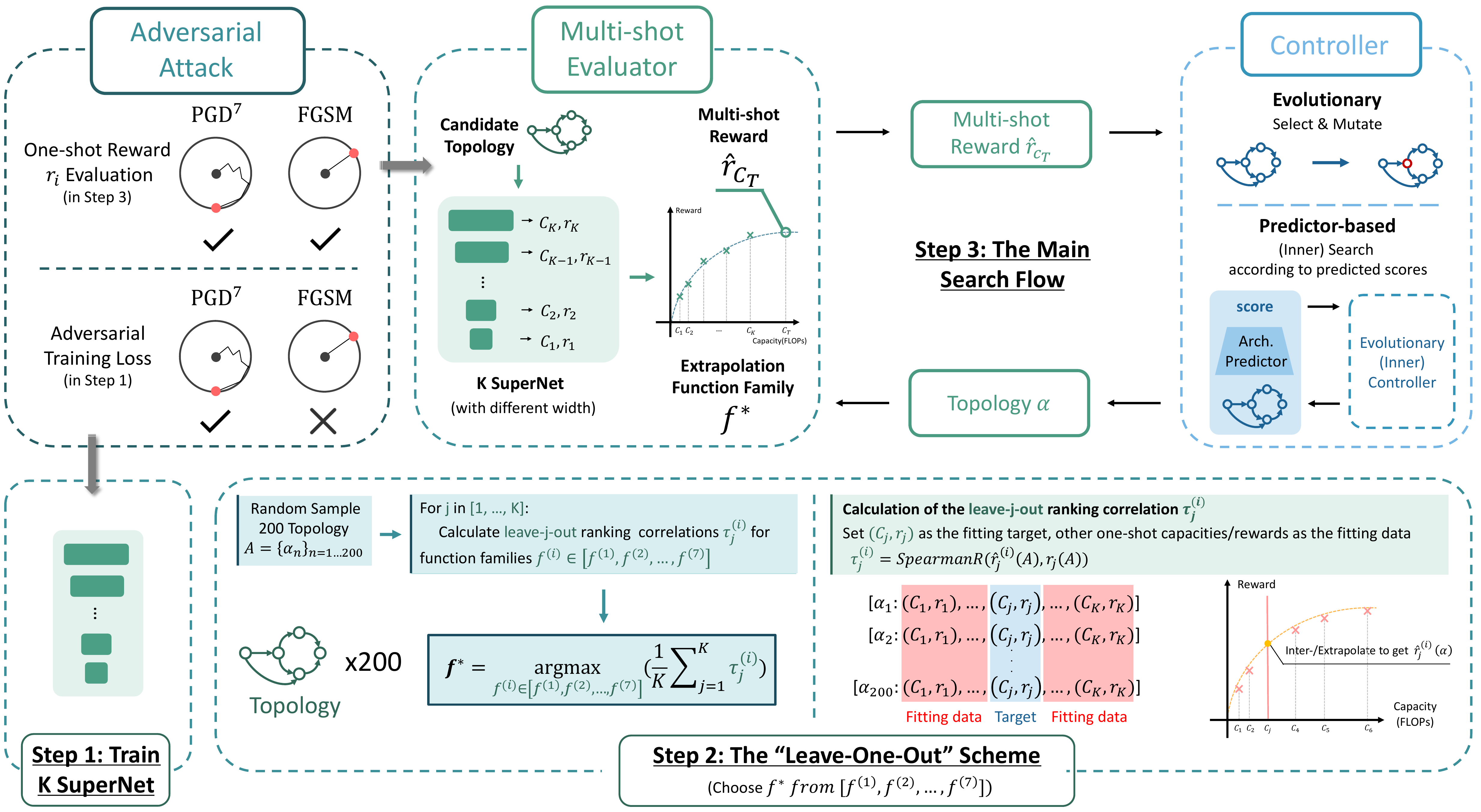}
  \caption{The overall workflow. Step 1: Adversarially train $K$ supernets with PGD$^7$ attack. Step 2: Select extrapolation function family $f^*$ based on the average leave-one-out Spearman ranking correlation (Sec.~\ref{sec:method_select}). Step 3: Conduct architecture search for targeted capacity $C_T$ (Sec.~\ref{sec:method_search}) using the multi-shot evaluation strategy (Sec.~\ref{sec:method_ms}).}
  \label{fig:flow} 
\end{figure*}

\section{Problem Definition}
\label{sec:pd}

The problem of searching for an adversarially robust architecture at the targeted capacity $C_{T}$ can be formalized as 
\begin{equation}
  \begin{aligned}
    \mbox{max}_{\alpha \in \mathcal{A}}& \mbox{  }r(\alpha)\\
    \mbox{s.t. }
     r(\alpha) &= E_{x_v \sim D_v} [ \mbox{min}_{x' \in B_p(x_v, \epsilon)}\\
    &\quad R(x', \mbox{Net}(\mbox{Aug}(\alpha, c(\alpha)), w^*(\alpha)))]\\
    c(\alpha) &= \argmin_{c \in \mathbb{Z}+} |C(\mbox{Aug}(\alpha, c)) - C_{T}|\\
    w^*(\alpha) &= \mbox{argmin}_{w} \loss_t(\alpha, w) \\
    &= \mbox{argmin}_{w} E_{x_t \sim D_t} [\mbox{max}_{x' \in B_p(x_t, \epsilon)} \\
    &\quad\quad \loss(x', \mbox{Net}(\mbox{Aug}(\alpha, c(\alpha)), w))],
     \end{aligned}
  \label{eq:robnas_formalization}
\end{equation}
where model augmentation in the channel (i.e., width) dimension is explicitly included to align architecture capacities to the target $C_T$: $\mbox{Aug}(\alpha, c)$ denotes the augmented architecture of topology $\alpha$ with init channel number $c$; $C(\mbox{Aug}(\alpha, c))$ refers to the capacity of the augmented architecture;
$c(\alpha)$ is an positive integer found by minimizing the difference of the augmented architecture capacity and the targeted capacity $C_{T}$. $B_p(x, \epsilon)$ denotes the $\epsilon$-ball around $x$ under $\ell_p$ norm, and both the reward $r(\alpha)$ and the loss $\loss_t(\alpha, w)$ are calculated in a min-max fashion. Compared with Eq.~\ref{eq:nas_formalization}, the search reward $r(\alpha)$ changes from $E_{x_v \sim D_v} [R(x_v, \mbox{Net}(\alpha, w^*(\alpha)))]$ to $E_{x_v \sim D_v}[\min_{x' \in B_p(x_v, \epsilon)} R(x', \mbox{Net}(\mbox{Aug}(\alpha, c(\alpha)), w^*(\alpha)))]$, where the two most notable changes are 1) The min-max formulation of adversarial robustness~\cite{madry2018towards}; 2) Model augmentation applied for each candidate to get an augmented architecture with a capacity $C(\mbox{Aug}(\alpha, c(\alpha))) \approx C_{T}$. 

In this paper, we propose a multi-shot strategy that can take advantage of the parameter sharing technique to accelerate the evaluation in this ``NAS for targeted capacity'' problem. This work adopts FLOPs to measure model capacity. Nevertheless, the multi-shot evaluation method can be extended to other capacity measures.



\section{Multi-shot NAS for Adversarial Robustness}\label{sec:multi_shot}

The overall workflow is illustrated in Fig.~\ref{fig:flow}. 
The multi-shot NAS workflow consists of three steps: 1) Supernet training: Adversarially train $K$ supernet $\chi_1, \cdots, \chi_K$ with init channels $c^s_1, \cdots, c^s_K$, and the corresponding shared weights are $w_1, \cdots, w_K$. A 7-step projected gradient descent attack (PGD$^7$)~\cite{madry2018towards} is used for the adversarial training as an approximation of the min-max optimization problem of $w_i$. 2) Function family selection: Select a extrapolation function family $f^*$ (Sec.~\ref{sec:method_select}). 3) Architecture search: Explore the search space (Sec.~\ref{sec:method_search}), and for each candidate topology, the multi-shot evaluation strategy is used to estimate its reward at the targeted capacity $C_{T}$ (Sec.~\ref{sec:method_ms}).

\subsection{Multi-shot Evaluation Strategy}\label{sec:method_ms}

To estimate the reward of a topology $\alpha$ at the targeted capacity $C_{T}$, $\alpha$ is firstly evaluated in all $K$ supernets to get its one-shot rewards $R(\alpha) = \{r_1(\alpha), \cdots, r_K(\alpha)\}$. 
Then, these one-shot rewards together with the corresponding one-shot capacity $C(\alpha) = \{C_1(\alpha), \cdots, C_K(\alpha)\}$ are used to fit the extrapolation function family $r = f(C; \beta)$, where $\beta$ is the parameters.
After getting $\hat{\beta} = \mbox{Fit}(f, R(\alpha), C(\alpha))$ with nonlinear least square fitting, the fitted function is used to estimate the reward $\hat{r}$ at the targeted capacity $C_{T}$.
\begin{equation}
  \hat{r}_{C_{T}} = f(C_{T}; \hat{\beta}).
\end{equation}

As we know, one-shot evaluation is the approximated evaluation strategy for the vanilla NAS problem (Eq.~\ref{eq:nas_formalization}): Instead of finding $w^*(\alpha)$ for each topology $\alpha$, the shared weights in the supernet are used to get the approximate one-shot reward. Correspondingly, multi-shot evaluation is the approximated evaluation strategy for the ``\textbf{NAS at targeted capacity}'' problem (Eq.~\ref{eq:robnas_formalization}): Instead of actually constructing the augmented network $\mbox{Net}(\mbox{Aug}(\alpha, c(\alpha)), \cdot)$ at the targeted capacity $C_{T}$ and training its weights to get $w^*(\alpha)$, we estimate the targeted reward by inter- or extra-polating multiple one-shot rewards.

Note that our method can capture the increasing tendency of each architecture's performance as the capacity increases. In this way, the search process takes the influence brought by model augmentation into consideration, instead of only treating the model augmentation as a post-processing step~\cite{hu2020tf,darts,pham2018efficient}. 

\noindent\textbf{One-shot Evaluation and BatchNorm Calibration } During each one-shot evaluation, we run PGD$^7$ or FGSM attacks on the validation dataset split and average the clean and adversarial accuracy as the reward $r_i(\alpha)$. Unlike previous studies~\cite{guo2019meets} that train each sub-architecture separately for several epochs, we use the shared weights for sub-architecture evaluation. However, there exist problems with batch normalization (BatchNorm): During the supernet training process, the accumulation of BatchNorm statistics is incorrect for each architecture. Thus, we calibrate BatchNorm statistics using the first 10 batches in the validation split.

\subsection{Extrapolation Function Family Selection}\label{sec:method_select}

Empirical observation~\cite{madry2018towards} shows that as the model capacity gets larger, the increase of the model robustness saturates. Due to this observation, we choose 7 parametric saturating function families and list them in the appendix. After training $K$ supernets, we select the appropriate extrapolation function family $f^*$ from $F=\{f^{(1)},\cdots,f^{(7)}\}$ by calculating the average leave-one-out (LOO) ranking correlation, as also illustrated in Fig.~\ref{fig:flow}.
Specifically, to assess the function family $f^{(i)}$, the leave-$j$-out ranking correlation $\tau_j^{(i)}$ for each supernet $\chi_j$ is calculated as follows. For each topology $\alpha$, we leave its score in $\chi_j$ out and fit $f^{(i)}$ to get $\beta_j^{(i)}(\alpha)=\mbox{Fit}(f^{(i)}, \{r_m(\alpha)\}_{m \in M}, \{C_m(\alpha)\}_{m \in M})$, and $M=\{m|m\in \{1,\cdots, K\}, m\neq j\}$. Then we can get the estimated reward of $\alpha$ in $\chi_j$: $\hat{r}_j^{(i)}(\alpha) = f^{(i)}(C_j(\alpha); \beta_j^{(i)}(\alpha))$.
Denoting $\hat{r}_j^{(i)}$ and $r_j$ for $N$ topologies as $\bm{\hat{r}_j^{(i)}}, \bm{r_j} \in \mathbb{R}^N$, the Spearman ranking correlation (SpearmanR) between the estimated and actual one-shot rewards of the topologies in $\chi_j$ is
\begin{equation}  \label{eq:corr_loo}
  \begin{aligned}
    \tau_j^{(i)} = \mbox{SpearmanR}&(\bm{\hat{r}_j^{(i)}}, \bm{r_j}),
  \end{aligned}
\end{equation}
where $N=200$ randomly sampled topologies are used in our experiments. After calculating $\{\tau_j^{(i)}\}_{i=1,\cdots,7, j=1,\cdots,K}$, we choose the function family with the highest average LOO SpearmanR $\bar{\tau}^{(i)}$ as the extrapolation family:
\begin{equation}
  \begin{aligned}
    f^* &= \argmax_{f^{(i)}\in F} \bar{\tau}^{(i)} = \argmax_{f^{(i)}\in F} \sum_{j=1}^K \tau_j^{(i)}.
  \end{aligned}
  \label{eq:average_loo}
\end{equation}

\subsection{Search with Multi-shot Evaluation Strategy}\label{sec:method_search}

After supernet training and function family selection, we explore the search space using the rewards evaluated by the multi-shot evaluation strategy (Sec.~\ref{sec:method_ms}). Two different search strategies (controllers) are used: 1) Evolutionary search; 2) Predictor-based search.

Note that since the supernet training phase and the architecture search phase is decoupled, after the supernets are trained only once, they can be easily reused in various architecture search processes with different capacity measures (e.g., \#Params, latency) and targeting different capacities.

\subsubsection{Search Space}\label{sec:method_ss}
\begin{figure}[tb]
  \centering
  \includegraphics[width=0.46\textwidth]{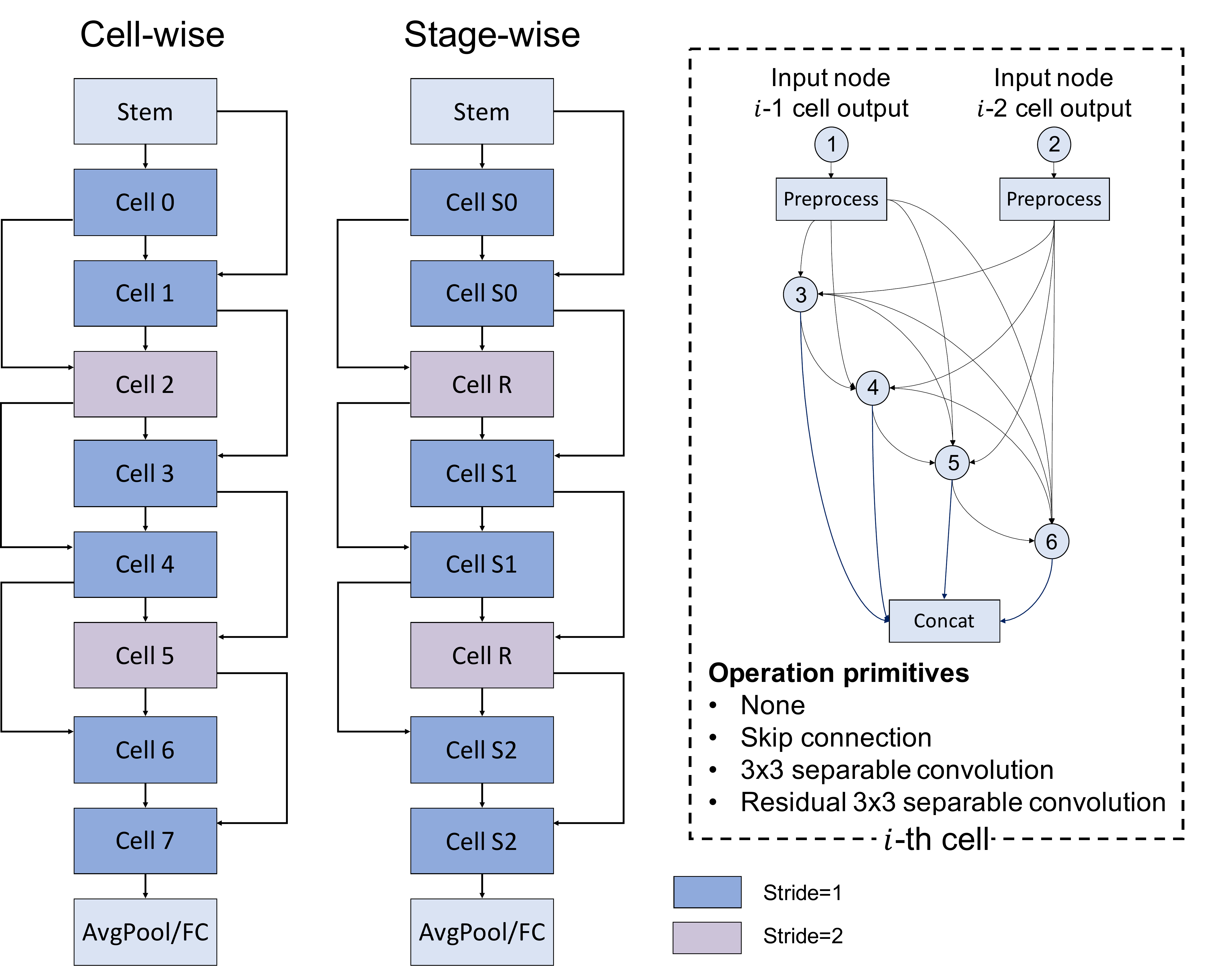}
  \caption{Illustration of the search space. Left: The layout and connection between cells (cell-wise/stage-wise). Right: Densely-connected cell search space.} 
  \label{fig:search_space} 
\end{figure}

Fig.~\ref{fig:search_space} summarizes the cell-level search space design and the cell layout. Our cell-level search space design is identical to \cite{guo2019meets}: There are four internal nodes, and each internal node can choose an arbitrary number of previous nodes as its inputs to enable discovering densely connected patterns.
Four primitives are included: none, skip connection, 3x3 separable convolution, residual 3x3 separable convolution. 

We experiment with two types of cell layout: cell-wise and stage-wise. In the cell-wise layout, each of the eight cells has a distinct topology ($2.7\times10^{67}$ architectures in the search space). In the stage-wise layout, the cells in each stage share the same topology, and the search space size is enormously reduced to $5.2\times10^{33}$. 


\subsubsection{Evolutionary Search}

We use the tournament-based evolutionary search strategy~\cite{real2019regularized} (population size 100, tournament size 10) to explore the cell-wise search space.  


In the previous NAS for robustness study~\cite{guo2019meets}, each candidate architecture is finetuned for $3$ epochs during the evaluation, which brings large computational costs.
In contrast, we eliminate the separate training costs during the search process by using the BatchNorm calibration technique (Sec.~\ref{sec:method_ms}).
By analyzing the search process (see the appendix for concrete analyses), we can see that as more and more architectures are explored, the relative speedup of multi-shot NAS to the NAS method with separate training phases increases and approaches $25/8\approx 3\times$.

\subsubsection{Predictor-based Search}
\label{sec:predictor-base-method}

\noindent\textbf{Predictor-based search strategy (controller)}:
The multi-shot evaluation strategy utilizes the parameter sharing technique and the extrapolation scheme to avoid separate augmentation and training costs for each architecture.
To further reduce the costs of separate inferences on the validation data, we propose to train an efficient architecture performance predictor and utilize this predictor to ``pick out'' promising architectures. In this way, only these promising architectures need to be evaluated on the validation data.

Specifically, we use a graph-based encoder~\cite{gates} to encode the topology of each cell into a continuous embedding, and then concatenate the embeddings of the four stage-wise cell topologies (S0, S1, S2, R) as the architecture embedding. Then the architecture embedding is fed into an MLP to get a predicted score. 

The proposed predictor-based flow goes as follows. We randomly sample 200 initial architectures and evaluate them with the multi-shot evaluator. Then, the architecture-performance pairs are used to train the initial predictor. After that, we run a predictor-based search for $8$ stages. In each stage, $2.5k$ architectures are fed into the predictor, and $50$ promising ones are picked out to be evaluated with the multi-shot evaluator. The evaluation results are then used to tune the predictor. The predictor training settings and how we pick out promising architectures (i.e., the inner search process) are elaborated in the appendix.

In our predictor-based search flow, $200 + 400 \times 50 \approx 20k$ architectures are assessed by the predictor, and only 3\% of these topologies ($200+400=600$) are picked and assessed with the multi-shot evaluation strategy.

\noindent\textbf{Stage-wise search space}:
With limited architectural information available for learning, predictors can provide better predictions in a smaller search space.
And during the experiment, we find that the cellwise search space is indeed too large for the initial predictor to make meaningful predictions.
Therefore, we only conduct the predictor-based search flow in the much smaller stage-wise search space, in which the predictor can be effectively trained. 
On the other hand, a smaller search space is easier to explore, which also facilitates a sufficient exploration in a limited search budget.
It is worthy to note that since the stage-wise search space is a sub-search space of the cell-wise one, the supernets could be directly reused and no extra training is needed.

\begin{figure}[tb]
  \centering
  \includegraphics[width=0.35\textwidth]{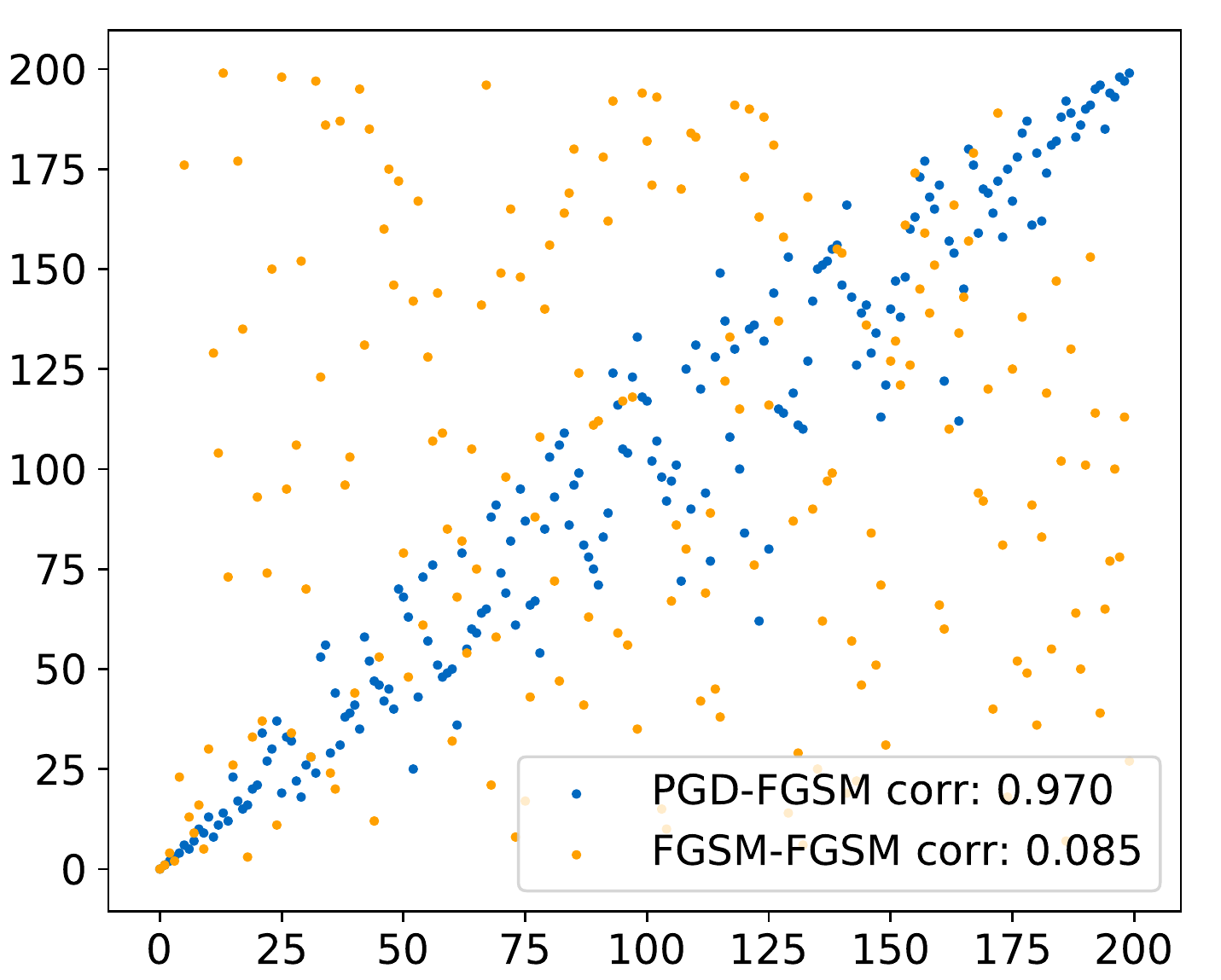}
  \caption{X-axis: Indices of architectures sorted according to reward ranking evaluated by PGD$^7$-PGD$^7$. Y-axis: The reward rankings evaluated by PGD$^7$-FGSM and FGSM-FGSM. The legends show the Spearman correlation between the rewards and the PGD$^7$-PGD$^7$ rewards. The supernet's init channel number is 24.}
  \label{fig:fgsm_compare_24c}
\end{figure}

\subsubsection{FGSM Reward as the Evaluation Proxy}
\label{sec:method_fgsm}
It is slow to evaluate the robustness of a model accurately, and this brings a large computational burden to the NAS process where thousands of architectures need to be evaluated. For example, using PGD$^7$ attack to measure the model robustness is almost as costly as $8$ clean model tests. Fortunately, NAS does not necessarily require the reward to be absolutely accurate, as long as the reward can reveal the relative ranking of architectures~\cite{gates}.
Thus, to alleviate the computational burden, we propose to use the FGSM attack as an efficient evaluation proxy of the PGD$^7$ attack to measure the adversarial accuracy. Moreover, we only calculate the reward on the first half of the validation split.

We verify the rationality of using FGSM as the proxy reward in Fig.~\ref{fig:fgsm_compare_24c}. The notation ``ATTACK1-ATTACK2'' means that ``ATTACK1'' is used in supernet adversarial training, and ``ATTACK2'' is used in the parameter-sharing evaluation. We can see that as long as a stronger attack (PGD$^7$) is used in adversarial training to avoid gradient masking, the FGSM rewards are highly correlated with the PGD rewards on 200 randomly sampled topologies (SpearmanR=0.97). 
The correlation analyses of more robustness criteria are discussed in Appendix D.

Note that a previous study on NAS for robustness~\cite{Chen2020AntiBanditNA} proposes to use FGSM attack during the whole search phase for efficiency. However, Fig.~\ref{fig:fgsm_compare_24c} illustrates that using a much weaker attack FGSM for adversarial training causes the supernet evaluation to be completely uncorrelated (SpearmanR=0.085). 

By virtue of using FGSM proxy reward and fewer validation images, we accelerate each multi-shot evaluation by roughly 8$\times$: From 25 to 3 minutes to evaluate each topology on a 2080Ti GPU. That is to say, while providing better reward estimation at targeted capacities (see Fig.~\ref{fig:function_selection}), each multi-shot evaluation could be finished in the same amount of time as a vanilla one-shot evaluation.

\section{Experiments}
\label{sec:exp}
We conduct multi-shot NAS on CIFAR-10, and evaluate the discovered architectures on the CIFAR-10, CIFAR-100, SVHN and Tiny-ImageNet~\cite{Krizhevsky2009LearningML,Netzer2011ReadingDI} datasets.

\begin{table*}[tb]
  \centering
  \caption{Comparison with baseline architectures under various adversarial attacks on CIFAR-10. MSRobNet-$C_T$ indicates the search is targeting at FLOPs $C_T$. The ``-P'' suffix indicates stage-wise architectures discovered by predictor-based search (Sec.~\ref{sec:results_pred}).} 
  \label{tab:cifar_manual_compare}
  \resizebox{\textwidth}{!}{
      \begin{tabular}{c|cc|ccccccc}
    \toprule
    \multirow{2}{*}{\textbf{Architecture}} &\multirow{1}{*}{\textbf{\begin{tabular}[c]{@{}c@{}}\#Param\\(M)\end{tabular}}} & \multirow{1}{*}{\textbf{\begin{tabular}[c]{@{}c@{}}\#FLOPs\\(M)\end{tabular}}}
    &\multicolumn{5}{c}{\textbf{Accuracies (\%)}} & \multicolumn{2}{c}{\textbf{Distances (median $\times 10^{-2}$)}}\\\cmidrule(lr){4-8}\cmidrule(lr){9-10}
    & &  & \textbf{Clean}&\textbf{FGSM}&\textbf{PGD$^7$}&\textbf{PGD$^{20}$} &\textbf{PGD$^{100}$}&\textbf{DeepFool}&\multicolumn{1}{c}{\textbf{BIM}} \\
        \midrule
            MobileNet-V2                             & 2.30        &182          &77.0       &53.0       &50.1       &48.0       &47.8            &3.07              &3.03\\
            VGG-16                                   &14.73        &626          &79.9       &53.7       &50.4       &48.1       &47.9       &3.25               &3.08\\
            ResNet-18                                &11.17        &1110         &83.9       &57.9       &54.5       &51.9       &51.5        &3.98 &3.41\\
             \textbf{MSRobNet-1000}                           &3.16        &1018          &\textbf{84.5$\pm$0.4}       &\textbf{59.6$\pm$0.2}       &\textbf{55.7$\pm$0.2}      &\textbf{52.7$\pm$0.3}       &\textbf{52.3$\pm$0.4}             &\textbf{4.08$\pm$0.05}          &\textbf{3.51$\pm$0.03}  \\
    \midrule
    RobNet-free~\cite{guo2019meets}   & 5.49        &1560         &    82.8         &    58.4         &     55.1        &   52.7         &   52.6\%                 &3.67          &3.47\\
    \textbf{MSRobNet-1560}    &5.30        &1588         &\textbf{85.1$\pm$0.2}       &\textbf{60.4$\pm$0.3}       &\textbf{56.4$\pm$0.4}       &\textbf{53.4$\pm$0.3}       &\textbf{53.1$\pm$0.3}                   &\textbf{4.34$\pm$0.05} &\textbf{3.60$\pm$0.02}\\
    \textbf{MSRobNet-1560-P}    &4.88         &1565         &85.0$\pm$0.2       &59.5$\pm$0.3 &55.6$\pm$0.3 &52.7$\pm$0.7 &52.3$\pm$0.7 &4.21$\pm$0.09 &3.52$\pm$0.04 \\
    \midrule
    RobNet-large~\cite{guo2019meets}  & 6.89        &2060  &78.6  &55.0   & - &49.4  &49.2 &-&- \\
    RobNet-large-v2~\cite{guo2019meets} &33.42      &10189  &85.7   &57.2  & - &50.5  &50.3 &-&-\\
    \textbf{MSRobNet-2000}                  &6.46         &2009        &\textbf{85.7$\pm$0.3}       &60.6$\pm$0.3       &\textbf{56.6$\pm$0.3}       &\textbf{53.6$\pm$0.4}      &\textbf{53.2$\pm$0.5}               &4.33$\pm$0.10          &3.60$\pm$0.02\\
    \textbf{MSRobNet-2000-P}                  &5.74         &2034        &85.7$\pm$0.3   & \textbf{60.8$\pm$0.4}      &56.5$\pm$0.3    &53.5$\pm$0.3      &53.1$\pm$0.3              &\textbf{4.49$\pm$0.06}             &\textbf{3.60$\pm$0.03}\\
    \bottomrule
    \end{tabular}%
  }
\end{table*}%

  \begin{table*}[tb]
  \centering
      \caption{Black-box PGD$^{100}$ attack accuracy on CIFAR-10. Adversarial examples are crafted on an independently trained substitute model and then used to attack the target model. ``MSRobNet'' is abbreviated as ``MSRN''.}

  \resizebox{\textwidth}{!}{
    \begin{tabular}{c|ccc|cccc}

      \toprule
      {\small \textbf{Target/Substitute}}            &{\small VGG-16} & {\small ResNet-18}   &{\small MobileNet-V2} & {\small MSRN-1560} & {\small MSRN-2000} &{\small MSRN-1560-P} &{\small MSRN-2000-P}\\\midrule
      {\small VGG-16}    &56.7\% &58.6\% &60.0\%  &59.7\% &59.9\%&59.8\% &60.2\%\\
      {\small ResNet-18} &62.5\% &59.3\%  &64.7\%&61.1\%&61.3\%&61.3\% &61.5\% \\
      {\small MobileNet-V2}     &57.8\% &58.0\% &54.7\% &58.8\% &59.0\% &58.7\% &59.0\%\\
      \midrule
      {\small MSRN-1560} &66.2\% & 63.4\%  &67.3\% &61.5\% &62.2\% &62.4\% &61.6\%\\
      {\small MSRN-2000} &66.6\% & 64.1\%  &68.0\%&62.8\% &61.6\%&62.9\% &62.7\%\\
      {\small MSRN-1560-P} &65.6\% &62.6\%  &67.0\%&62.2\%&61.8\%&61.8\% &62.0\%\\
      {\small MSRN-2000-P} &67.3\% & 64.7\% &68.6\% &63.1\%&63.2\%&63.3\% &62.0\%\\
      \bottomrule
    \end{tabular}
  }
    \label{tab:blackbox}
  \end{table*}%
  
\subsection{Experimental Setup}

\subsubsection{Search Settings}
During the search process, the original training dataset of CIFAR-10 is divided into two parts: training split (40000 images) and validation split (10000 images). The supernets are trained on the training split, and architecture rewards are evaluated on the validation split.
PGD$^7$ under $\ell_\infty$ norm with $\epsilon$=$0.031 (8/255)$ and step size $\eta$=$0.0078 (2/255)$ is used for adversarial training.
We train $K=8$ supernets with init channel number $\{12, 24, 30, 36, 40, 44, 54, 64\}$ for 400 epochs. 
We use a batch size of 64, a weight decay of 1e-4, and an SGD optimizer with a momentum of 0.9. And the learning rate is set to 0.05 initially and decayed down to 0 following a cosine schedule.

After training $K$ supernets, we select the extrapolation function family $f^*$ with the described method in Sec.~\ref{sec:method_select}.
Then, we use $f^*$ in the search for robust architectures at various targeted FLOPs: 1000M, 1560M, 2000M.



\subsubsection{Adversarial Training and Testing}
For the final comparison on CIFAR-10, CIFAR-100 and SVHN, we adversarially train the architectures for 110 epochs using PGD$^7$ attacks with $\epsilon=0.031 (8/255)$ and step size $\eta=0.0078 (2/255)$, and other settings are also kept the same. Detailed settings are elaborated in the appendix. 

To evaluate the adversarial robustness of the trained models, we apply the Fast Gradient Sign Method (FGSM)~\cite{Goodfellow2015ExplainingAH} with $\epsilon=0.031 (8/255)$ (4/255 for Tiny-ImageNet) and PGD~\cite{madry2018towards} with different step numbers.
On CIFAR-10, we also apply another two attacks that report the successful attacking $\ell_{\infty}$ distances using the Foolbox toolbox V2.4.0~\cite{rauber2017foolbox} and its default settings, and report the median attacking distance on the test dataset:
1) DeepFool~\cite{moosavi2016deepfool}; 2) Basic Iterative Method (BIM)~\cite{Kurakin2017AdversarialEI}.

\subsection{Results on CIFAR-10}

Tab.~\ref{tab:cifar_manual_compare} compares the performances of the architectures under various adversarial attacks. The architectures discovered by our method are referred to as MSRobNet-$C_T$, in which $C_T$ is the targeted FLOPs in multi-shot search. We train MSRobNets for 4 times with different seeds, and report the mean and standard deviation of their performances.
As the baseline architectures, we choose several manually designed architectures, and we also compare with recent NAS-discovered ones~\cite{guo2019meets} in a search space that is similar to ours.
We can see that with similar FLOPs, the architectures discovered by multi-shot NAS significantly outperform the baseline architectures.
For example, at the targeted FLOPs of 2000M, our discovered architectures, MSRobNet-2000 and MSRobNet-2000-P, surpass RobNet-large by a large margin of 4\%-7\% with fewer FLOPs and much fewer parameters.
Also, at the targeted FLOPs of 1560M, MSRobNet-1560 (clean 85.1\%, PGD$^{7}$ 56.4\%) outperforms the recent NAS-discovered architecture RobNet-free (clean 82.8\%, PGD$^{7}$ 55.1\%) significantly. 

In addition, we also run two baseline \textit{one-shot} search methods. Denoting $r^{(A)}(\alpha) = \frac{1}{2} (\mbox{acc}_{\mbox{adv}} + \mbox{acc}_{\mbox{clean}}), r^{(B)} = \mbox{FLOPs}(\alpha)$, the two baseline one-shot search methods are:
1) One-shot-rA: Search with reward $r^{(A)}(\alpha)$; 2) One-shot-div: Search with a scalarized reward $r(\alpha) = r^{(A)}(\alpha)/ r^{(B)}(\alpha)$ that takes both the performance and capacity into consideration. The comparison of MSRobNets with the architectures discovered by baseline one-shot methods are presented in the appendix.\\


\begin{table*}[!h]
    \centering
    \caption{One-shot rewards \& FLOPs and multi-shot rewards of the stage-wise architectures discovered by predictor-based search. }

    \begin{tabular}{cc|cccccc}
      \toprule
      \multirow{2}{*}{\textbf{Topology}}     & \multirow{2}{*}{\textbf{\begin{tabular}[c]{@{}c@{}}Search\\Reward\end{tabular}}}      & \multicolumn{4}{c}{\textbf{One-shot Rewards / FLOPs (M)}} & \multicolumn{2}{c}{\textbf{Multi-shot Rewards}} \\
      \cmidrule(lr){3-6}\cmidrule(lr){7-8}
      & & $r_2$ (24c) & $r_6$ (44c) & $r_7$ (54c) & $r_8$ (64c) & $\hat{r}_{1560}$ & $\hat{r}_{2000}$\\\midrule
      one-shot-rA-24c-P & $r_2$ (24c) & \textbf{0.594} / 329 & 0.623 / 1037 & 0.630 / 1539 & 0.636 / 2140 & 0.631 & 0.634 \\
      one-shot-rA-44c-P & $r_6$ (44c) & 0.580 / 340 & \textbf{0.624} / 1072 & 0.639 / 1592 & 0.642 / 2213 & 0.635 & 0.642 \\
      \midrule
      MSRobNet-1560-P & $\hat{r}_{1560}$ & 0.569 / \textbf{253} & 0.619 / \textbf{805} & 0.638 / \textbf{1196} & 0.645 / \textbf{1666} & 0.645 & 0.652 \\
      MSRobNet-2000-P & $\hat{r}_{2000}$ & 0.572 / 261 & 0.620 / 826 & \textbf{0.646} / 1228 & \textbf{0.648} / 1708 & \textbf{0.649} & \textbf{0.660} \\
      \bottomrule
      \end{tabular}
    \label{tab:predictor-based-ablation}
  \end{table*}

    \begin{table}[tb]
    \centering
        \caption{Accuracy comparison on CIFAR-100.} 
    \resizebox{\linewidth}{!}{
    \begin{tabular}{c|c|cccc}
      \toprule
      \textbf{Architecture}  &\textbf{\begin{tabular}[c]{@{}c@{}}\#FLOPs(M)\end{tabular}} &\textbf{Clean}  & \textbf{PGD$^{20}$} & \textbf{PGD$^{100}$} \\\midrule
      VGG-16           &  626          &51.5\%              &25.8\% &25.8\% \\            
      ResNet-18        &  1100         &59.2\%              &29.9\% &29.7\% \\
      MobileNet-V2     &  182          &48.2\%              &26.3\% &26.2\% \\
      RobNet-free~\cite{guo2019meets} & 1560 &57.7\%  &31.1\% &30.8\% \\ \midrule
      MSRobNet-1000   &  1019 &60.1\% &31.2\% &31.1\% \\
      MSRobNet-1560   &  1588 & 60.8\%   & \textbf{31.7\%} & \textbf{31.5\%} \\
       MSRobNet-2000   & 2009 & \textbf{61.6\%} & 31.6\% & \textbf{31.5\%}\\
      \bottomrule
      \end{tabular}%
    }
    \label{tab:transfer}%
  \end{table}%

\begin{figure}[tb]
  \centering
  \includegraphics[width=0.90\linewidth]{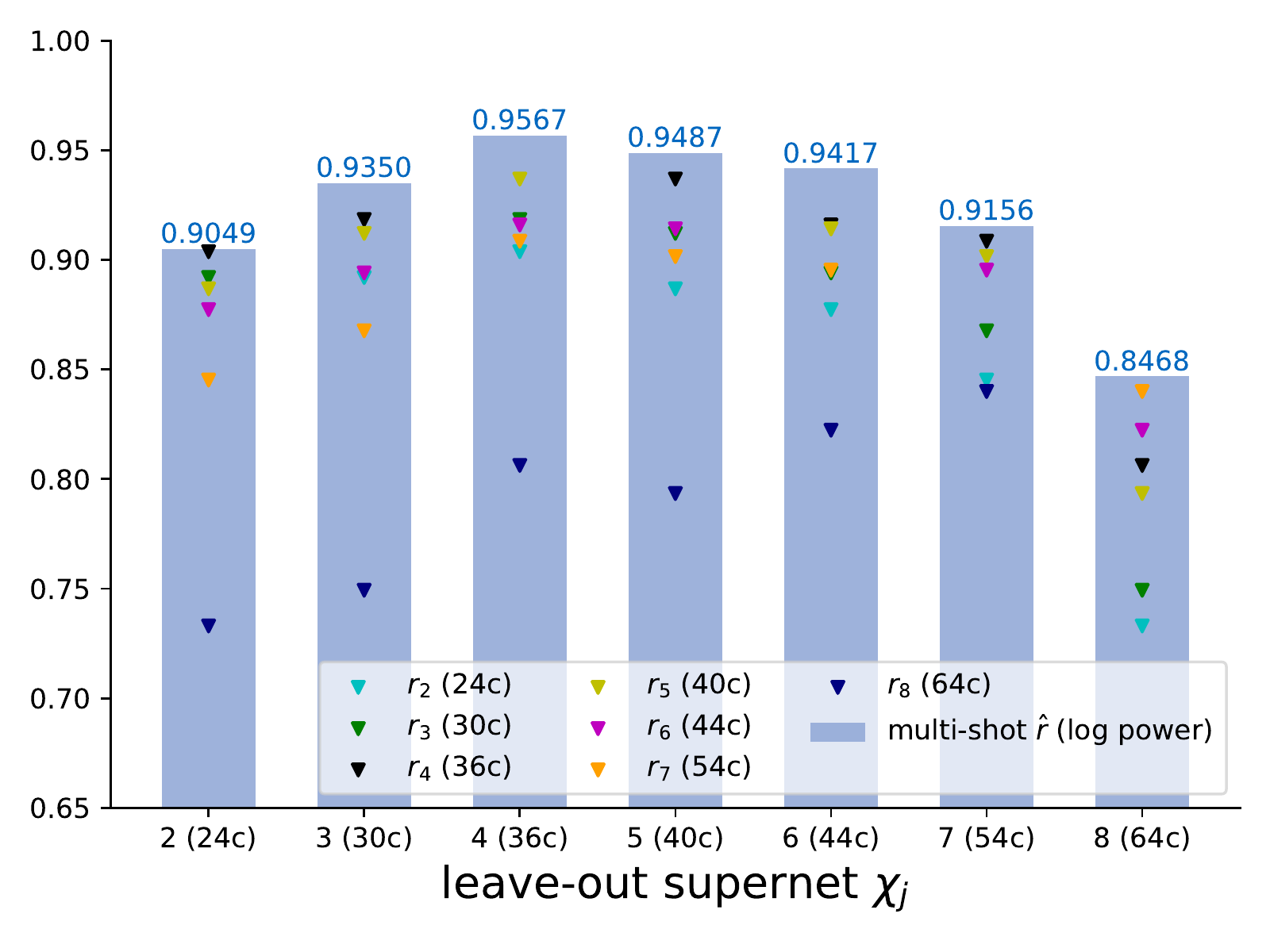}
  \caption{Spearman ranking correlation with the rewards $\bm{r_j}\in\mathbb{R}^{200}$ in supernet $\chi_j$. Bar: The LOO correlation $\tau_j = \mbox{SpearmanR}(\bm{\hat{r}}, \bm{r_j})$ in Eq.~\ref{eq:corr_loo}; Triangles: The correlations $\mbox{SpearmanR}(\bm{r_k}, \bm{r_j}), k\neq j$ between the one-shot rewards in other supernets and $\bm{r_j}$.}
  \label{fig:function_selection} 
\end{figure}

 \noindent\textbf{Black-box Attacks }
 Tab.~\ref{tab:blackbox} shows the results of transfer-based black-box evaluation \cite{papernot2016transferability}. 
 Specifically, we test the target model with PGD$^{100}$ adversarial examples crafted on the substitute model and report the accuracy. To evaluate the adversarial transferability between models with the same architecture, we train another model of this architecture as the substitute model with the same training settings. We can see that compared to transferring between the baseline architectures, it is harder to transfer the attacks from baseline architectures to MSRobNets.\\


\subsection{Transfer Results to Other Datasets}
We train the architectures on CIFAR-100, SVHN, and Tiny-ImageNet. Tab.~\ref{tab:transfer} and SVHN/Tiny-ImageNet results in the appendix show that MSRobNets outperform the baselines significantly with comparable FLOPs.



  \subsection{More Efficient Predictor-based Search}
  \label{sec:results_pred}


  We conduct the predictor-based search using the multi-shot rewards targeting two FLOPs (1560M, 2000M), and the discovered topologies are named MSRobNet-1560-P and MSRobNet-2000-P, respectively. 
  Their final performances on CIFAR-10 are reported in Tab.~\ref{tab:cifar_manual_compare}. The whole predictor-based search process only takes 1.3 GPU days.

We also conduct the predictor-based search using two baseline one-shot-rA rewards: $r_2, r_6$, which are the one-shot rewards in the supernets with 24 and 44 init channels, respectively.
  Tab.~\ref{tab:predictor-based-ablation} shows the one-shot rewards / FLOPs and multi-shot rewards of topologies discovered with different search rewards. We can see that as indicated by the motivating illustration in Fig.~\ref{fig:motivating}, one-shot search prefers topologies with larger one-shot capacity, while these topologies are no longer superior when augmented to a different capacity. For example, the topology \textit{one-shot-rA-24c-P} discovered using reward $r_2$ has the highest $r_2=0.594$ and a high FLOPs of $329$M with 24 init channels. However, in the 64-init-channel supernet, it only achieves $r_8=0.636$ with 2140M FLOPs, while MSRobNet-2000-P achieves $r_8=0.648$ with a smaller FLOPs of 1708M. That is to say, although \textit{one-shot-rA-24c-p} is superior to MSRobNet-2000-P when their init channel number is 24, MSRobNet-2000-P surpasses \textit{one-shot-rA-24c-p} when they are augmented to use 64 init channels (FLOPs around 2000M). 

\subsection{Function Family Selection}

We choose ``log power'' as the extrapolation function family, and show the LOO ranking correlations of the 7 candidate function families in the appendix.
Fig.~\ref{fig:function_selection} shows that for each leave-out supernet $\chi_j$, the LOO ranking correlation $\tau_j$ of the multi-shot estimated rewards (blue bar) is higher than the correlations between $\bm{r_j}$ and the one-shot rewards in other differently-sized supernets (triangles).
This indicates that the multi-shot strategy effectively captures the effect brought by width expansion, and bridges the correlation gap brought by width difference.


\section{Conclusion}\label{sec:conc}


This paper proposes a multi-shot neural architecture search (multi-shot NAS) framework to discover adversarially robust architectures \textbf{at targeted capacities}. Instead of using one supernet in one-shot NAS, multiple supernets with different capacities are constructed. By evaluating each topology with the inter- or extra-polation of multiple one-shot rewards, our method can explicitly search for superior architectures at targeted capacities.
Experimental results demonstrate the effectiveness of the proposed method.

This work applies multi-shot NAS in the context of adversarial robustness, since the trade-off of capacity and adversarial robustness is very significant. Interesting directions for future work include: 1) Extending the multi-shot method to other application scenarios with a targeted capacity. 2) Sharing parameters between differently-sized supernets to reduce the supernet training cost.

\clearpage
{\small
\bibliographystyle{ieee_fullname}
\bibliography{egbib}
}

\clearpage
  \renewcommand{\thefigure}{A\arabic{figure}}
\setcounter{figure}{0}
\renewcommand{\thetable}{A\arabic{table}}
\setcounter{table}{0}

\begin{appendices}
  \section{Extrapolation Function Family Selection}
The 7 parametric saturating function families are
\begin{enumerate}
\item Janoschek: $ f(x; a,b,c,d) = a-(a-b)e^{-cx^d}$
\item vapor pressure: $ f(x; a,b,c) = e^{a + \frac{b}{x} + c\log(x)}$
\item log log linear: $ f(x; a,b) = \log(a\log(x) + b)$
\item ilog$_2$: $ f(x; a,c) = c - \frac{a}{\log(x)}$
\item log power: $ f(x; a,b,c) = \frac{a}{1+(x/e^b)^c}$
\item MMF: $ f(x; a,b,c,d) = a - \frac{a-b}{1+(dx)^c}$
\item log power rep: $ f(x; a, b, c) = \frac{1}{(1+e^{-a})(1+e^c x^{e^{-b}})}$ 
\end{enumerate}

 The leave-one-out function selection scheme is summarized in Alg.~\ref{alg:Leave-one-out-function-selection}. In practice, we find that all the candidate function families have a weak correlation between estimated and actual one-shot rewards in $\chi_1$ (supernet with init channel number 12). In addition, the capacity of $\chi_1$ is so small that FLOPs of almost all the topologies in it are far from the targeted capacity.
 Thus, we regard the estimation results in $\chi_1$ as outliers, and do not include $\tau_1^{(i)}$ when calculating $\bar{\tau}^{(i)} = \frac{1}{K-1} \sum_{j=2}^K \tau_j^{(i)}$.

\begin{figure}[h]
  \centering
  \includegraphics[width=\linewidth]{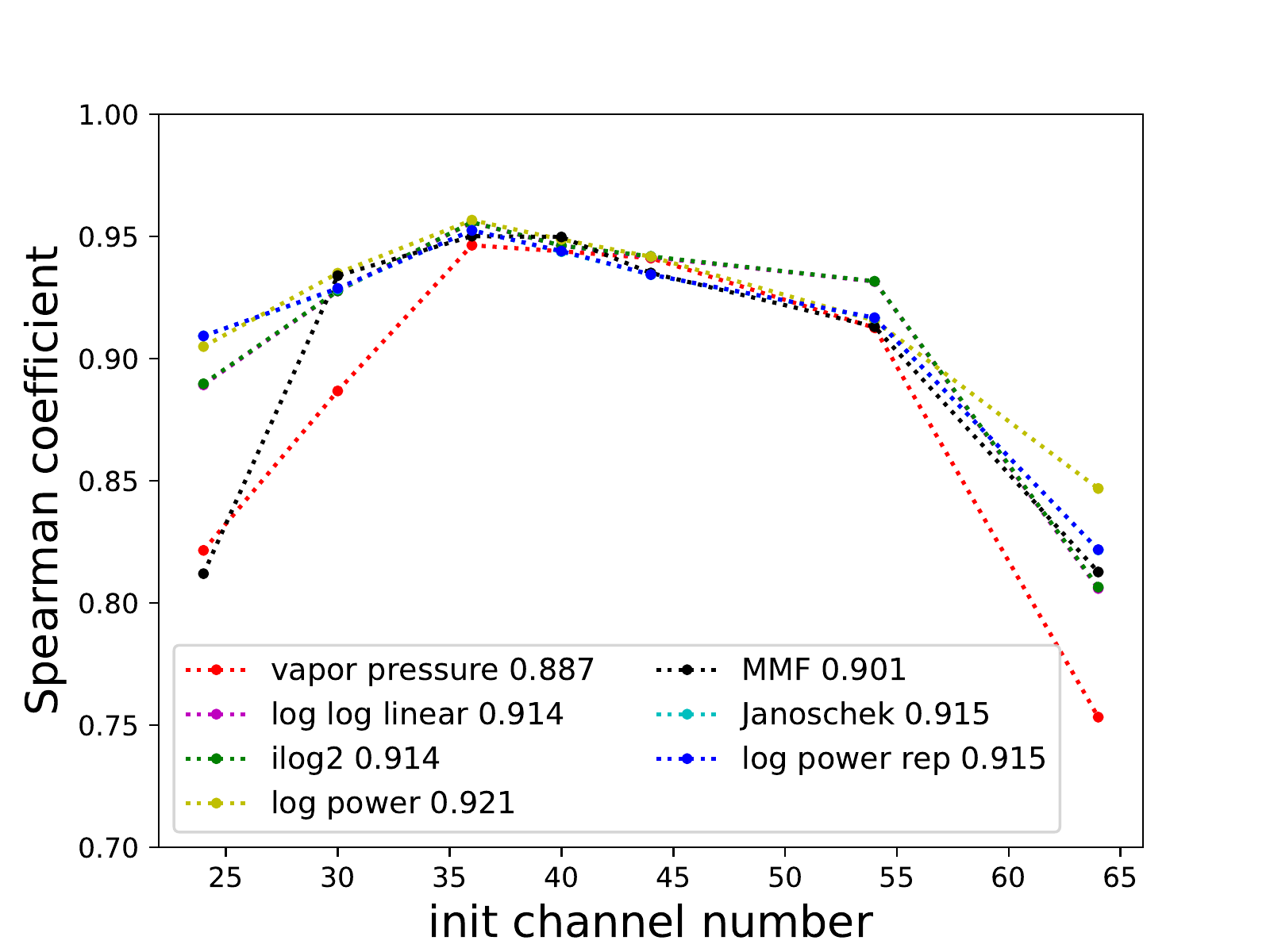}
  \caption{The leave-one-out Spearman ranking correlation curves of 7 function families (200 random sampled toplogies are used). Their average leave-one-out correlation coefficients $\bar{\tau}^{(i)}$s are shown in the legend.}
   \label{fig:function_selection_app} 
 \end{figure}
 
The leave-one-out correlations of 7 function families are shown in Fig.~\ref{fig:function_selection_app}. We choose ``log power'' as the extrapolation function family, since it has the highest average leave-one-out correlation (0.921).

\begin{figure}[tb]
  \centering
  \includegraphics[width=\linewidth]{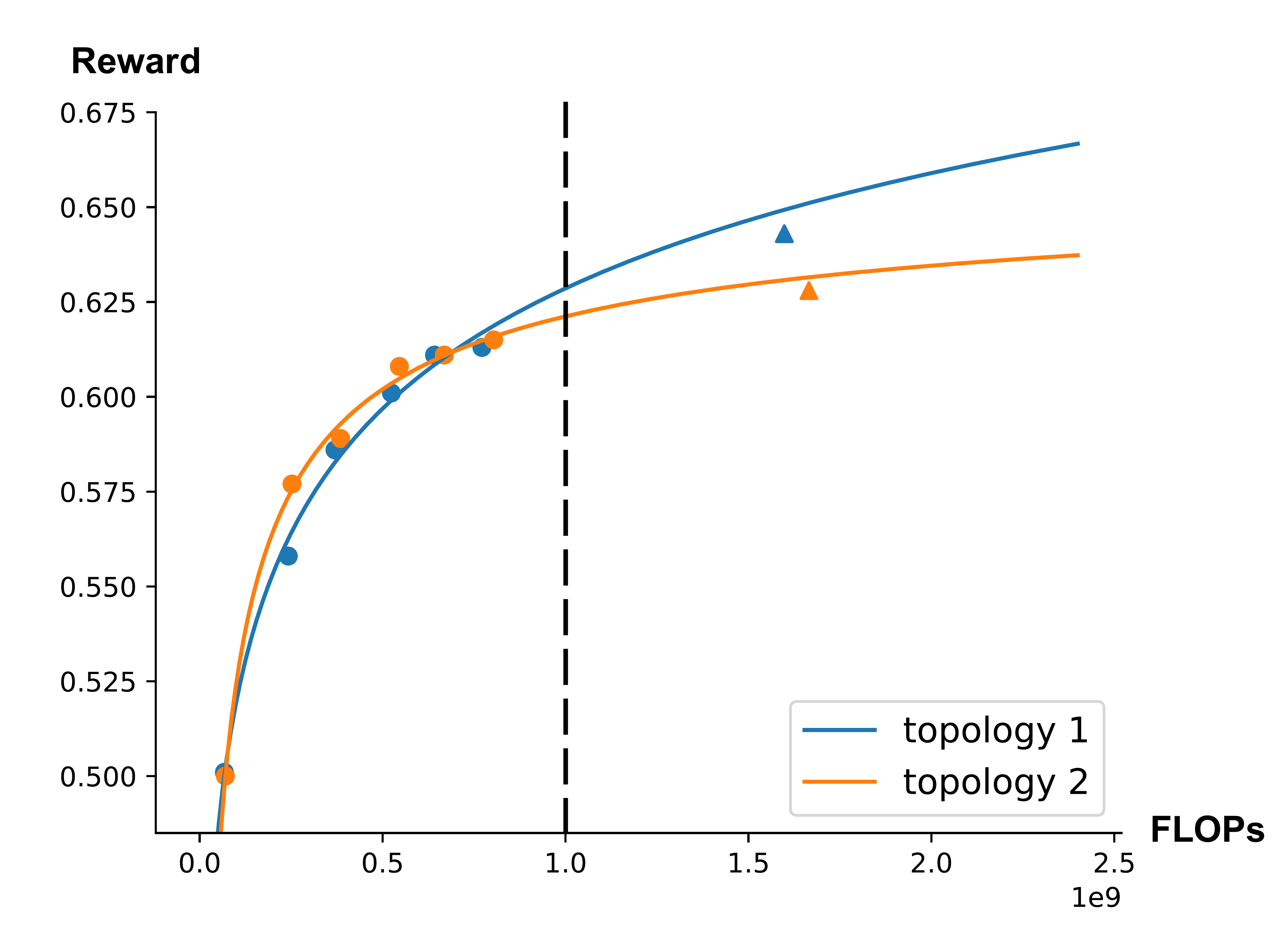}
  \caption{Reward extrapolation example. The markers on one curve denote the topology's rewards evaluated in differently-sized supernets.}
  \label{fig:inter}
\end{figure}

Tab. 3 in the main text gives some evidence on our motivation. To give a more intuitive evidence on the motivating example in the main text, we show the extrapolation curves 
of two topologies (among the 200 randomly sampled topologies in our experiment) in Fig.~\ref{fig:inter}.
We can see that when aligned to the same capacity (FLOPs), topology 2 is more robust in the low-capacity region, while topology 1 is more robust in the high-capacity region. 
The two curves are fitted using the data points at the left of the dashed line. And we can see that the fitted curve can capture the tendency, thus manage to predict the relative ranking correctly at a larger capacity without actual training a large supernet (the leftmost triangular markers).

\begin{algorithm*}
    \vspace{5pt}
  \caption{Multi-shot Architecture Search for Adversarial Robustness}
  \label{alg:multi_shot}
  \LinesNumbered
   \KwIn{
    \\
    \textit{- Data \& Attack}: train-valid split ratio $s=0.8$, PGD attack steps $T^a=7$, perturbation budget $\epsilon=0.031$, attack step size $\eta=0.0078$
    \\\textit{- Supernet training}: number of supernets $K=8$, supernet init channel numbers $\{c^s_1, \cdots, c^s_K\}$, training epochs $T^s=400$
    \\\textit{- Function Family Selection}: number of randomly sampled toplogies $N=200$, candidate function families $F=\{f^{(1)}, \cdots f^{(7)}\}$
    \\\textit{- Search}: targeted capacity $C_{T}$
    \begin{itemize}
    \item Evolutionary: search steps $T^e= 200$, population size $\pi=100$, tournament size $\mu=10$, PGD attack steps $T^{sa}=7$
    \item Predictor-based: search stages $T^p=8$, topologies sampled per stage $N^p=50$, inner evolutionary search steps $T^{pe}=N^p\times 50=2500$, PGD attack steps $T^{sa}=1$ (FGSM, no random init)
    \end{itemize}
   }
  \KwOut{A discovered topology.}
\nonl  $\empty$ \\
  $s$ of the training dataset as the training split $D_t$, reset as the validation split $D_v$\\
  \nonl  /* Adversarially train the supernets */\\
  \For{$i=1 \cdots K$}{
    $\chi_i, w_i \leftarrow \mbox{INIT-SUPERNET}(c^s_i)$\\
    \For{$j=1 \cdots T^s$}{
      \For{$x_t, y_t$ in $D_t$}{
        $\alpha \sim \mathcal{A}$ \quad/* random arch. sample */\\
        $x' \leftarrow \mbox{PGD}(x_t, y_t, \chi_i(\alpha, w_i); T^a, \epsilon, \eta)$\\
        $L = \mbox{CE}(\chi_i(\alpha, w_i)(x'), y_t)$\\
        $w_i \leftarrow \mbox{SGD-UPDATE}(w_i, \nabla_{w_i} L)$\\
      }
    }
  }
  \nonl $\empty$\\
  Select the extrapolation function family $f\in F$ based on the leave-one-out ranking correlations, using $N$ randomly sampled topologies\\
 \nonl $\empty$\\
  \If{EVO Search}{
    \nonl  /* Evolutionary Search with multi-shot evaluation */\\
    $\mbox{pop} \leftarrow$ \{$\pi$ topologies with the highest rewards in the $N$ random sampled topologies, together with their multi-shot estimated rewards at $C_T$\}\\
    \For{$t=1 \cdots T^e$}{
      $\mbox{pool} \leftarrow$  set of $\mu$ randomly choosed topologies from the $\mbox{pop}$\\
      $\mbox{parent} \leftarrow$ topologies with the best reward in $\mbox{pool}$\\
      $\alpha_t \leftarrow \mbox{MUTATION}(\mbox{parent})$\\
      Add $(\alpha_t, \mbox{MultiShotEval}(\alpha_t, C_T, f, T^{sa}))$ to $\mbox{pop}$\\
      Remove the worst topology in $\mbox{pop}$\\
    }
  }
  \ElseIf{Predictor-based search}{
    \nonl  /* Predictor-based Search with multi-shot evaluation */\\
    $\mbox{pop} \leftarrow$ \{the $N$ random sampled topologies, together with their multi-shot estimated rewards at $C_T$\}\\
    Train the intial $P$ using $\mbox{pop}$\\
    \For{$t=1 \cdots T^p$}{
      $\{\alpha^{(t)}_i\}_{i=1,\cdots,N^P} \sim$ evolutionary inner search for $T^{pe}$ steps using the predictor $P$\\
      Get their multi-shot estimated rewards $\{(\alpha^{(t)}_i, \mbox{MultiShotEval}(\alpha^{(t)}_i, C_T, f, T^{sa}))\}_{i=1,\cdots,N^P}$, and add to pop\\
      Tune $P$ with the newly evaluated topologies\\
    }
  }
  {\bfseries Return} topology with highest multi-shot estimated reward in the population $\mbox{pop}$
\end{algorithm*}

\begin{algorithm*}
  \caption{Leave-One-Out Function Selection}
  \label{alg:Leave-one-out-function-selection}
  \LinesNumbered
  \KwIn{Candidate Function Families $F=\{f^{(i)}\}_{i=1,\cdots,7}$}
  \KwOut{A function family $f^*$}
  \Parameter{Number of supernets $K=8$, number of topologies $N=200$}
  \nonl  $\empty$ \\
  Random sample $N$ architectures $\{\alpha_n\}_{n=1,\cdots,N}$ from the search space\\
\For{$j=1 \cdots K$}{
  Get the one-shot capacities and rewards $\{(C_j(\alpha_n), r_j(\alpha_n))\}_{n=1,\cdots,N}$\\
}
\For{$j=1 \cdots K$}{
  \nonl  /* Get leave-$j$ out correlation */\\
  \For{$i=1 \cdots 7$}{
    \For{$n=1 \cdots N$}{
      $\beta_j^{(i)}(\alpha) = \mbox{Fit}(f^{(i)}, \{r_m(\alpha)\}_{m=1, \cdots, K, m \neq j}, \{C_m(\alpha)\}_{m=1, \cdots, K, m \neq j})$\\
      $\hat{r}_j^{(i)}(\alpha_n) = f^{(i)}(C_j(\alpha_n); \beta_j^{(i)}(\alpha_n))$\\
    }
    $\tau_j^{(i)} = \mbox{SpearmanR}(\{\hat{r}_j^{(i)}(\alpha_n)\}_{n=1,\cdots,N}, \{r_j(\alpha_n)\}_{n=1,\cdots,N})$\\
  }
}
$f^* = \argmax_{f^{(i)}\in F} \bar{\tau}^{(i)} = \argmax_{f^{(i)}\in F} \sum_{j={\bf 2}}^K \tau_j^{(i)}$\\

  {\bfseries Return} function family $f^*$.
\end{algorithm*}

\begin{algorithm*}
  \caption{\textbf{MultiShotEval}: Multi-shot Architecture Evaluation}
  \label{alg:multi_shot_eval}
  \LinesNumbered
  \KwIn{A candidate topology $\alpha_t$, targeted capacity $C_T$, extrapolation function family $f$, PGD attack steps in adversarial accuracy evaluation $T^{sa}$}
  \KwOut{The multi-shot estimated reward $\hat{r}(\alpha_t)$ at capacity $C_T$}
  \Parameter{Accuracy coefficient $\lambda=0.5$}
\nonl  $\empty$ \\
      \For{$i=1 \cdots K$}{
        $w_i \leftarrow \mbox{CALIB-BN}(\alpha_t, w_i, D_v)$\\
        $r_i(\alpha_t) = (1-\lambda) \mbox{Acc}_{\mbox{clean}}(\chi_i(\alpha_t, w_i), D_v, y_v) + \lambda \mbox{Acc}_{\mbox{pgd}}(\chi_i(\alpha_t, w_i), D_v, y_v; T^{sa}, \epsilon, \eta)$\\
      }
      $\hat{\beta}(\alpha_t) = \mbox{Fit}(f, \{r_i(\alpha_t)\}_{i=1 \cdots K}, \{C_i(\alpha_t)\}_{i=1 \cdots K})$\\
      $\hat{r}(\alpha_t) = f(C_{T}; \hat{\beta}(\alpha_t))$\\
        {\bfseries Return} $\hat{r}(\alpha_t)$
    \end{algorithm*}

\section{Experimental Setup}
\subsection{Search Strategy Settings}
\noindent\textbf{Evolutionary Search }
The search strategy is a tournament-based evolutionary method~\cite{real2019regularized}, with population size 100 and tournament size 10. In each step, the worst topology is removed from the population.

 \begin{table*}[tb]
   \centering
   \caption{Comparison with baseline architectures under various adversarial attacks on SVHN. We retrain and test RobNet-free with the same setting.}
   \begin{tabular}{c|cc|ccccc}
     \toprule
     \textbf{Architecture} &\textbf{\#Param (M)} & \textbf{\#FLOPs (M)} & \textbf{Clean}&\textbf{FGSM}&\textbf{PGD$^7$}&\textbf{PGD$^{20}$}&\textbf{PGD$^{100}$}\\
     \midrule
           VGG-16           & 14.73& 626          &92.3\%       &66.6\% &55.0\% &47.4\%       &45.1\% \\            
     ResNet-18        & 11.17& 1100         &92.3\%       &73.5\% &57.4\% &51.2\%       &48.8\% \\
     MobileNet-V2     & 2.30& 182            &93.9\%       &73.0\% &61.9\% &55.7\%       &53.9\% \\
     RobNet-free~\cite{guo2019meets} & 5.49& 1560  &94.2\% &84.0\% &66.1\% &59.7\% &56.9\%\\ \midrule
     MSRobNet-1000   & 3.16& 1019 &94.5\% & \textbf{85.5\%} &\textbf{67.8\%} & \textbf{61.6\%} & \textbf{58.6\%} \\
     MSRobNet-1560   & 5.30& 1588 & \textbf{95.0\%} & 77.5\% &64.0\% &57.0\% &54.2\%\\
      MSRobNet-2000   &6.56& 2009 &94.9\% & 84.8\% &65.3\% &58.8\% &55.1\%\\
     \bottomrule
     \end{tabular}%
   \label{tab:svhn_transfer}%
 \end{table*}%

    \begin{table*}[tb]
    \centering
    \caption{Comparison with baseline architectures under various adversarial attacks on Tiny-ImageNet.}
    \begin{tabular}{c|cc|ccccc}
      \toprule
      \textbf{Architecture} &\textbf{\#Param (M)} &\textbf{\#FLOPs (M)} &\textbf{Clean} &\textbf{FGSM} &\textbf{PGD$^5$} &\textbf{PGD$^{20}$} &\textbf{PGD$^{100}$}\\\midrule
          ResNet-18 &11.28& 1126 &50.3\% &27.1\% &28.8\% &24.4\% &24.3\%\\
          ResNet-50 &23.92& 2612 &52.2\% &27.8\% &29.4\% &25.2\% &25.0\%\\\hline
          MSRobNet-1000 &3.33& 1056 &52.3\% &\textbf{28.7\%} &\textbf{30.6\%} &\textbf{26.5\%} &\textbf{26.4\%}\\
          MSRobNet-1560 &5.51& 1633 &\textbf{52.9\%} &28.1\% &29.7\% &25.9\% &25.6\%\\
      \bottomrule
      \end{tabular}%
    \label{tab:tiny-transfer}
  \end{table*}

\noindent\textbf{Predictor-based Search }
The predictor-based search process is summarized in Alg.~\ref{alg:multi_shot}. We first randomly sample 200 stage-wise architectures, evaluate them, and train the initial predictor (Line 21). Then, the predictor-based search is run for $T^p=8$ stages. In each stage, $N^p=50$ architectures are sampled by the predictor-based controller (Line 23) and evaluated using the multi-shot strategy (Line 24). And then, the multi-shot evaluation results are used to tune the predictor (Line 25).

In the inner search process (Line 23), we run a tournament-based evolutionary search with population size 20 and tournament size 5. We conduct $N^p\times 50=25k$ evolutionary search steps, and in each step, a mutated topology is assessed by the predictor, which is very efficient. And in every 50 steps, a topology with the highest predicted score in the current population is decided for the actual multi-shot evaluation. In each step, the topology with the lowest predicted score is removed from the population.

As for the predictor construction, we adapt a recent graph-based encoder~\cite{gates} to our search space. Specifically, we encode the topology of each cell into a continuous vector, and then concatenate the embeddings of four cell topologies (S0, S1, S2, R) as the architecture embedding. Then the architecture embedding is fed into an MLP to get a predicted score. The operation embedding dimension, node embedding dimension, and hidden dimension are all 48. And the output dimension of each cell topology is set to 32. Thus, the dimension of one full topology is 128. Then this 128-dim vector is fed into a 3-layer MLP with 256 hidden units and output a final score. In each predictor training process, the predictor is trained for 100 epochs with batch size 50 and an Adam optimizer (1e-3 learning rate). We use hinge pair-wise ranking loss with margin $m=0.1$ to train the predictor following \cite{gates}. A dropout of 0.1 before the MLP is used.

\subsection{Adversarial Training and Testing of Architectures}
For the final comparison of all the architectures, we adversarially train them for 110 epochs on CIFAR-10/CIFAR-100 and 50 epochs on SVHN. As for the adversarial attack, we adopt the PGD$^7$ attack with $\epsilon=0.031 (8/255)$ and step size $\eta=0.0078 (2/255)$. The batch size is set to 48, and we use an SGD optimizer with momentum 0.9, gradient clipping 5.0, and weight decay 5e-4. 
The learning rate is set to 0.05 initially. On CIFAR-10 and CIFAR-100, the learning rate is decayed by 10 at epoch 100 and epoch 105. And on SVHN\footnote{To successfully train a VGG-16 model on SVHN, we first conduct a normal training process for 50 epochs, and finetune the normally-trained model with adversarial training.}, it is decayed by 10 at epoch 20, epoch 40 and epoch 45.
In addition, following a recent study~\cite{pang2020bag}, we employ the label smoothing regularization~\cite{Szegedy2016RethinkingTI} with weight 0.2. No cutout and dropout is applied. 
All models are trained with the same settings on a single 2080Ti GPU.  

On Tiny-ImageNet, the image size is 64$\times$64 instead of 32$\times$32, and we change the stem from a 3x3 convolution to a 7x7 convolution with stride 2. Due to this modification, the overall FLOPs is still close to $C_T$. For all the architectures, we first conduct a normal training process for 110 epochs with batch size 256. The learning rate is set to 0.1 initially and decayed by 10 at epoch 30, 60 and 100. Using the normally-trained model as the pre-trained model, we adversarially train the model for 70 epochs with PGD$^5$ attack ($\epsilon=0.0157(4/255)$, step size $\eta=0.0039(1/255)$). The learning rate is linearly warmed up from 0 to 0.05 in 10 epochs, and decayed by 10 at epoch 45 and 60.
In both the normal and adversarial training processes, an SGD optimizer with momentum 0.9, gradient clipping 5.0, and weight decay 3e-4 is used. 

The settings of the distance attacks on CIFAR-10: 1) DeepFool: 100 steps; 2) Basic Iterative Method (BIM): 20 binary search steps.

\section{Additional Experimental Results}

\subsection{BatchNorm Calibration}

\begin{figure}[tb]
  \centering
  \includegraphics[width=0.46\textwidth]{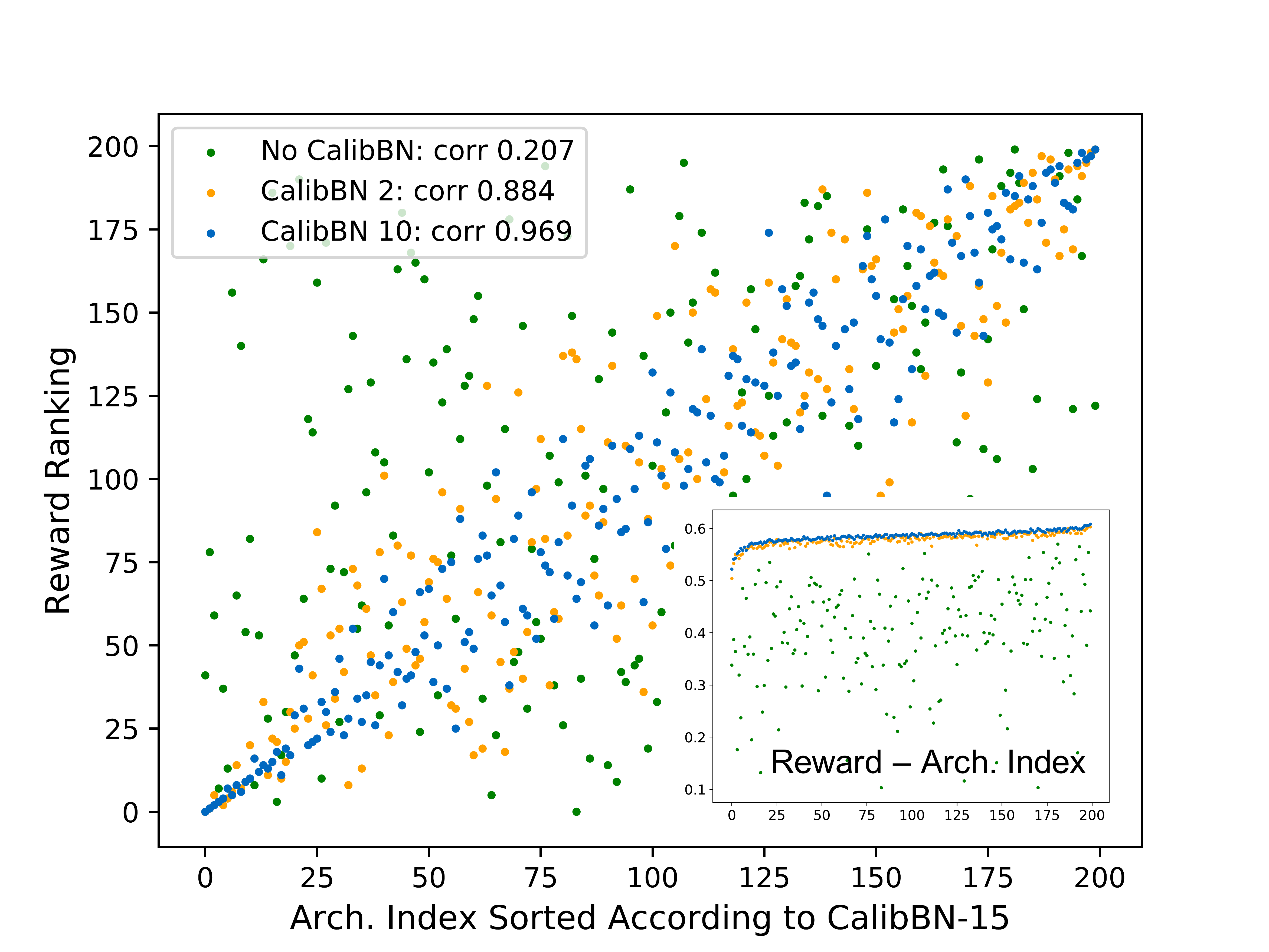}
  \caption{The effect of BatchNorm calibration.}
  \label{fig:plot_rank}
\end{figure}

Fig.~\ref{fig:plot_rank} shows that without BatchNorm calibration (No CalibBN, green points), the evaluation results are unmeaningful. And after BN calibration with several batches, the accuracies and rankings of 200 randomly sampled sub-architectures become meaningful. 
Note that we calculate the ranking correlation with the CalibBN-15 results (calibration using 15 batches), since we fail to tune the architectures to the same level of accuracies with only 3 epochs of training as in \cite{guo2019meets}.

\begin{table*}[tb]
  \centering
  \caption{Comparison with architectures discovered by baseline one-shot workflows on CIFAR-10. $C_T$ in ``One-shot-*-$C_T$'' indicates that the discovered architecture are augmented to have $C_T$M FLOPs. All the one-shot search are run in the 44-channel supernet.}
  \label{tab:cifar_oneshot_compare}
  \begin{tabular}{c|cc|ccccc}
    \toprule
    \textbf{Architecture}            &\textbf{\#Param (M)} &\textbf{\#FLOPs (M)}&\textbf{Clean}&\textbf{FGSM}&\textbf{PGD$^7$}&\textbf{PGD$^{20}$}&\textbf{PGD$^{100}$}\\
    \midrule
    One-shot-rA-1560               &5.65 &1571 &84.9\%  &60.0\% &56.0\%&53.1\%            &52.6\%\\
    One-shot-div-1560              &5.00 &1552 &84.1\% &59.1\%  &55.0\%&52.0\%       &51.6\% \\
    MSRobNet-1560                  &5.30         &1588         &\textbf{85.1\%}       &\textbf{60.4\%}      &\textbf{56.4\%}  & \textbf{53.4\%}      &\textbf{53.1\%}\\\midrule
    One-shot-rA-2000                &7.25 &2013 &85.3\% &59.9\% &55.9\% &53.0\% &52.7\%\\
    One-shot-div-2000              &6.46           &2005         &85.1\%       &59.8\%       &55.9\% & 52.8\%        &52.4\%\\
    MSRobNet-2000                  &6.46          &2009         &\textbf{85.7\%}       &\textbf{60.6\%}       &\textbf{56.6\%}      & \textbf{53.6\%}  &\textbf{53.2\%}\\
    \bottomrule
  \end{tabular}
\end{table*}

\subsection{Transfer Results to SVHN and Tiny-ImageNet}
We train MSRobNets and all baseline architectures with identical settings that have been described before, and show the comparison in Tab.~\ref{tab:svhn_transfer} and Tab.~\ref{tab:tiny-transfer}.
We can see MSRobNets outperform the baseline architectures, with comparable FLOPs and fewer parameters.

An interesting observation is that the architectures demonstrate slightly different relative robustness on these two datasets. For example, MSRobNet-1000 outperforms MSRobNet-1560 in adversarial accuracies on SVHN/Tiny-ImageNet, and RobNet-free outperforms MSRobNet-1560 on SVHN. This may arise from that larger dataset difference degrades the architecture transferability. Thus, the MSRobNet-1560 architecture discovered on CIFAR-10 is suboptimal on SVHN/Tiny-ImageNet for the targeted FLOPs of 1560M. %

\subsection{Comparison with Baseline One-shot Workflows}
We compare with the architectures discovered by baseline one-shot workflows at targeted FLOPs 1560M and 2000M in Tab.~\ref{tab:cifar_oneshot_compare}. All the one-shot searches are conducted on the supernet with init channel number 44. After the search, the discovered topology is augmented to the targeted capacity $C_{T}$.
We can see that MSRobNets indeed outperform those architectures discovered by the one-shot methods consistently.

\section{Discussion}

\begin{figure*}[tb]
  \centering
  \includegraphics[width=0.82\textwidth]{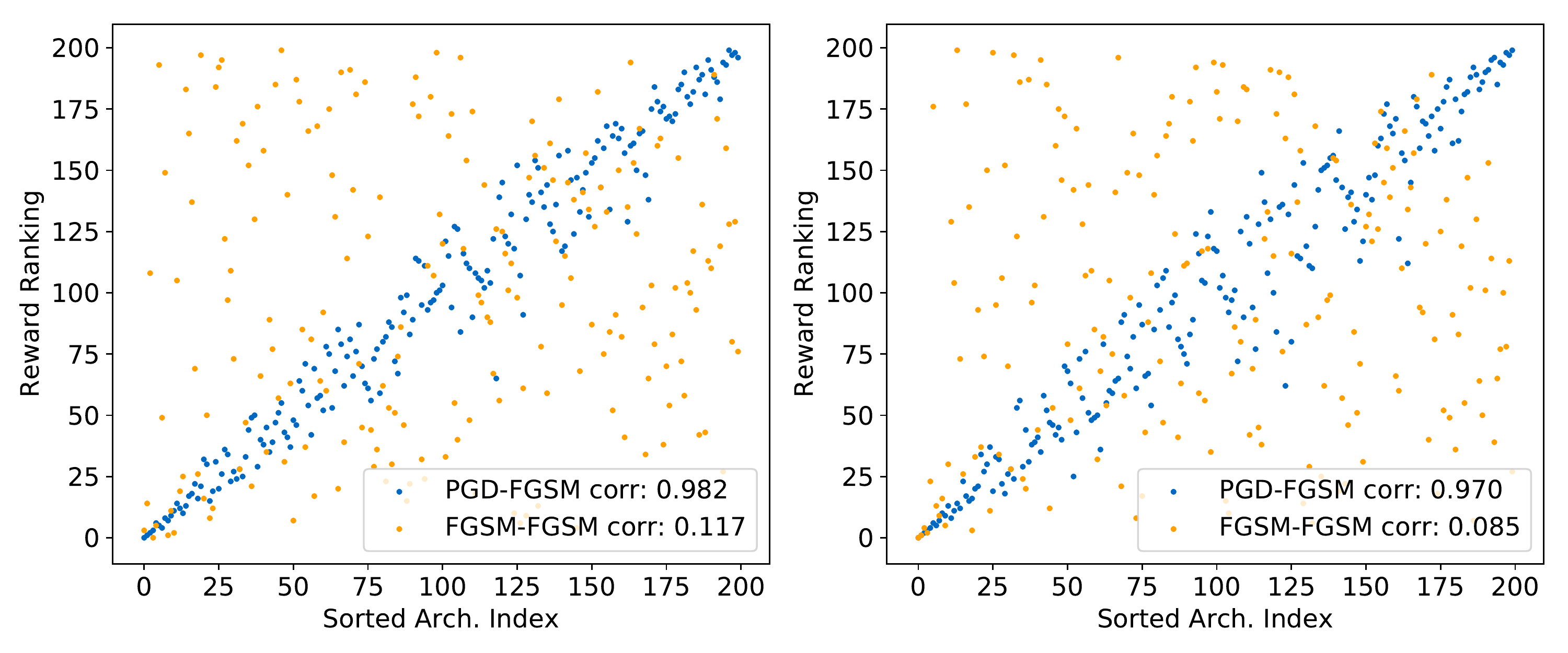}
  \caption{X axis: Indices of architectures sorted according to reward ranking evaluated by PGD$^7$-PGD$^7$. Y axis: The reward rankings evaluated by PGD$^7$-FGSM and FGSM-FGSM. The legends show the Spearman correlation of the rewards and the PGD$^7$-PGD$^7$ rewards. Left: 24 init channel; Right: 36 init channel. }
    \label{fig:fgsm_compare}
  \end{figure*}
  
\subsection{Adversarial Robustness Evaluation in the Search Phase: PGD V.S. FGSM}

Compared with normal training and evaluation, adversarial training and adversarial robustness evaluation are much more time-consuming. Since NAS itself faces the computational challenge, how to evaluate the adversarial robustness efficiently in the search phase is an important problem. We wonder whether we can use a weaker and faster attack in the search phase while still managing to discover a robust architecture under the stronger attack. For example, while the final comparison is conducted using PGD$^7$ (PGD attack with 7 steps), a recent study~\cite{Chen2020AntiBanditNA} tried to use the weaker FGSM attack during the whole search phase for efficiency.

As described in the main text, the search phase includes the supernet training phase and the parameter-sharing search phase.
We conduct the following experiments to verify whether it is suitable to use FGSM as a substitute in the search phase: 1) FGSM-FGSM: Train another supernet with FGSM adversarial training, and use FGSM to evaluate the 200 randomly sampled topologies. 2) PGD$^7$-FGSM: Reuse the supernet trained with PGD$^7$ adversarial training, but use FGSM to evaluate the 200 randomly sampled topologies. The notation ``ATTACK1-ATTACK2'' means that ``ATTACK1'' is used in supernet adversarial training and ``ATTACK2'' is used in the parameter-sharing evaluation, and the step size of the FGSM attack is set to 8/255. Fig.~\ref{fig:fgsm_compare} shows the rankings of these evaluations and the Spearman coefficient with the PGD$^7$-PGD$^7$ evaluations. The results show that using a much weaker attack FGSM for adversarial training causes the supernet evaluation to be uncorrelated. Thus, one can use FGSM for acceleration during the parameter-sharing search phase, while should keep using a stronger attack during the supernet training phase.

\begin{table*}[tb]
    \centering
    \caption{Kendall Tau, Spearman correlation, and P@K of using PGD$^7$-FGSM as a proxy reward for PGD$^7$-PGD$^7$ in different supernets. 200 randomly sampled stage-wise topologies are used.}
    \begin{tabular}{c|cc|ccccc}
      \toprule
      \textbf{Supernet}            &\textbf{Kendall Tau} & \textbf{Spearman} & \textbf{P@5} & \textbf{P@10} &\textbf{P@20} &\textbf{P@50} &\textbf{P@100}\\\midrule
      $\chi_1$    &0.904 &0.987 &0.60 &0.80 &0.90 &0.88 &0.94\\
      $\chi_2$    &0.862 &0.972 &0.80 &0.90 &0.75 &0.90 &0.92\\
      $\chi_3$    &0.859 &0.973 &0.60 &0.60 &0.80 &0.90 &0.94\\
      $\chi_4$    &0.832 &0.960 &0.80 &0.70 &0.90 &0.82 &0.94\\
      $\chi_5$    &0.867 &0.973 &0.60 &0.90 &0.85 &0.90 &0.93\\
      $\chi_6$    &0.858 &0.969 &0.60 &0.90 &0.75 &0.88 &0.91\\
      $\chi_7$    &0.835 &0.959 &0.80 &0.90 &0.70 &0.82 &0.93\\
      $\chi_8$    &0.830 &0.958 &0.80 &0.80 &0.75 &0.84 &0.90\\
      \bottomrule
      \end{tabular}
    \label{tab:proxycompare}
  \end{table*}%

  In the stage-wise search space, we also calculate the Kendall Tau, Spearman correlation, and P@K~\cite{gates} of using PGD$^7$-FGSM as a proxy reward for PGD$^7$-PGD$^7$. The P@K criterion reports the proportion of topologies with top-K rewards evaluated by PGD$^7$-FGSM in the top-K topologies evaluated by PGD$^7$-PGD$^7$, and is a criterion that focus more on well-performing topologies.
  Results in Tab.~\ref{tab:proxycompare} show that most of the high-score topologies evaluated by PGD$^7$-FGSM also perform well under PGD$^7$-PGD$^7$ evaluation, which demonstrates that FGSM is indeed a suitable proxy attack for PGD during the search phase.

\subsection{Adversarial Robustness Evaluation in the Search Phase: Distance-based Criterion}

In our paper, we use a reward formulation that is a weighted sum of clean and adversarial accuracies.

\begin{equation}
  \begin{aligned}
    r = \lambda acc_{adv} + (1-\lambda) acc_{clean},
  \end{aligned}
  \label{eq:reward_define}
\end{equation}
where $\lambda$ is a hyperparameter making a trade-off between the clean and adversarial accuracies. During our search, $\lambda$ is set to 0.5. In other words, we use the average of clean and adversarial accuracies as the reward to guide the search.

Another type of robustness criterion is the distance-based criteria. The Foolbox toolbox~\cite{rauber2017foolbox} implements various kinds of distance attacks that attempt to find a minimum successful perturbation. Examples that have been misclassified without any adversarial perturbation are assigned zero attack distance. In this way, no averaging coefficient is needed to trade off the clean and adversarial accuracies, since the clean accuracy is handled implicitly.

We try to explore how these different robustness criteria correlate with each other. We evaluate the 200 randomly sampled architectures on the validation dataset in $\chi_2$ (init channel number 24) under five kinds of distance attacks. The five attacks are Basic Iterative Method (BIM) $\ell_2$ / $\ell_\infty$~\cite{Kurakin2017AdversarialEI}, DeepFool $\ell_2$ / $\ell_\infty$~\cite{moosavi2016deepfool} and C\&W $\ell_2$~\cite{carlini2017towards}, where $\ell_2$ or $\ell_\infty$ denotes the type of the $\ell_p$ distance measure in the input space. Due to the low efficiency of C\&W attack, we only run it on 500 images ($1/20$) in the validation set. We use the median distance as one architecture's performance criterion under the preformed attack, since it is insensitive to outliers (e.g., when the attack fails to find a proper adversarial perturbation within limited attack steps). For comparison, we also run PGD$^7$ and FGSM attacks and calculate their rewards as in Eq.~\ref{eq:reward_define} with $\lambda=0.5$.

\begin{figure}[bt]
  \centering
  \includegraphics[width=\linewidth]{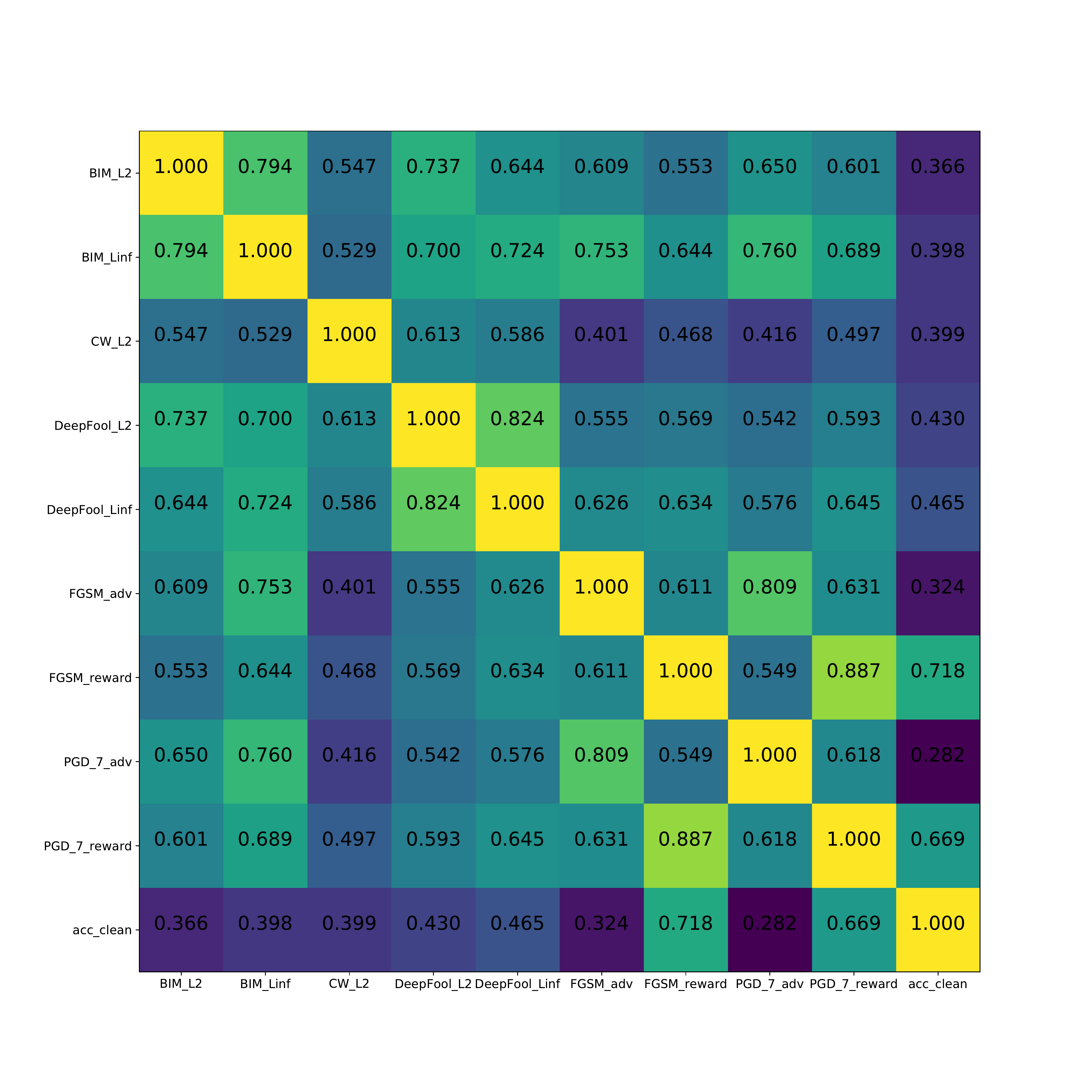}
  \caption{The correlation of various robustness criteria during the search phase (200 cell-wise topologies are used).}
   \label{fig:distance_corr} 
\end{figure}

The Kendall Tau correlation coefficients between different criteria are shown in Fig.~\ref{fig:distance_corr}. We can see that
\begin{enumerate}
\item Different robustness criteria are indeed correlated.
\item Compared with the relatively high correlation between FGSM and PGD adversarial accuracies and rewards ($> 0.8$), their correlations with distance criteria are relatively weaker. The clean accuracy is better correlated with FGSM/PGD rewards (0.718, 0.669) than the distance-based criteria (0.366-0.465). This is intuitive since the clean accuracy is added into FGSM/PGD rewards explicitly, while only handled implicitly (attack distance$=0$ or $>0$) in distance-based attacks. On the other hand, they are all higher than the correlations of clean accuracy and FGSM/PGD adversarial accuracies (0.324, 0.282), which verifies that an implicit or explicit trade-off between clean and adversarial accuracies must be considered in the robustness criterion.
  \item Intuitively, compared with $\ell_\infty$ attack distances, $\ell_2$ attack distances are better correlated with other $\ell_2$ attack distances, and vice versa. For example, $Corr(\mbox{DeepFool } \ell_{\infty}, \mbox{BIM } \ell_{\infty}) = 0.724 > Corr(\mbox{DeepFool } \ell_{\infty}, \mbox{BIM } \ell_{2}) = 0.644$. There are indeed differences brought by input space measure, and this motivates us to use criteria with different input space measure to assess model robustness.
  \item FGSM/PGD attacks are conducted with $\ell_\infty$ input space measure, thus the correlation of their rewards with $\ell_\infty$ attack distances is higher than $\ell_2$ attack distances.
  \end{enumerate}

\subsection{Efficiency}

As mentioned in the main text, the FGSM proxy reward and fewer validation images accelerate the evaluation process by roughly 8$\times$ (from 25 min to 3 min for evaluating each topology on a single 2080Ti GPU).

After been trained only once, the supernets can be reused to discover architectures targeting different capacities (e.g., 1000M, 1560M, 2000M), or in search space with different macro layouts (e.g., cell-wise, stage-wise). 
Thus, for each targeted capacity, denoting the epoch-wise training and evaluation time of one architecture as $t^s$ and $t^e$, the search time can be estimated as $(N+T^e) K t^e$.
By using the BatchNorm calibration technique, we obviate the need of separate architecture training phases during the search. If each architecture needs to be finetuned for $n$ epochs (e.g., $n=3$ in \cite{guo2019meets}), the search time becomes $(\pi+T^e) (t^e + nt^s)$. We can estimate the coefficients of $T^e$ in the multi-shot method and a NAS method with separate training phases as $K t^e \approx 8 t^e$ and $t^e + n t^s \approx 25 t^e$, respectively.\footnote{In our experiments, we find that $t^s/t^e$ is approximately $2$ times of the train-valid data portion ratio $t^s/t^e \approx 2 s/(1-s)= 8$.} This means that, compared with the NAS methods that have separate training phases, the relative efficiency of multi-shot NAS emerges as the number of explored architectures increases and the speedup would approaches 25/8 asymptotically.

In the predictor-based search flow, $N+T^p \times N^p=200+400=600$ topologies in total are assesed using the multi-shot evaluation strategy, and the whole search process can be finished in 1.3 GPU days.

\section{Discovered Architectures}

\begin{figure*}[tb]
  \centering
  \includegraphics[width=\textwidth]{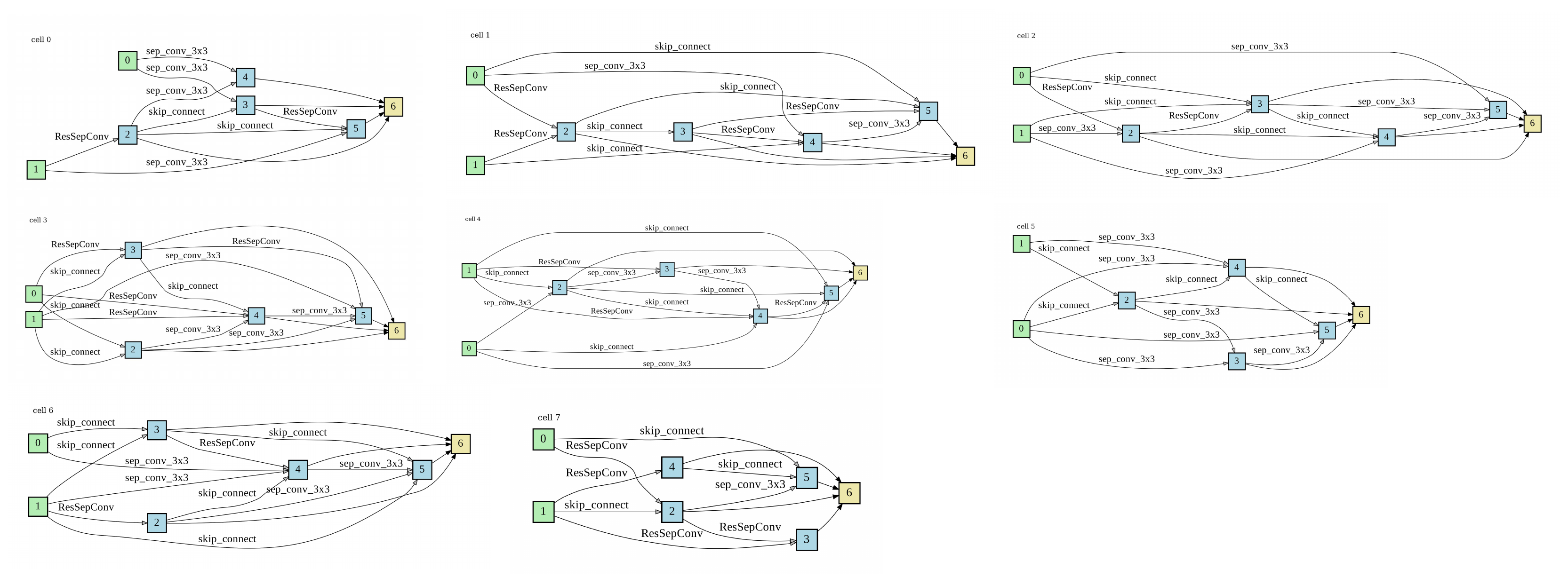}
  \caption{Architecture of MSRobNet-1000.}
  \label{fig:msrobnet-1000}
\end{figure*}

\begin{figure*}[tb]
  \centering
  \includegraphics[width=\textwidth]{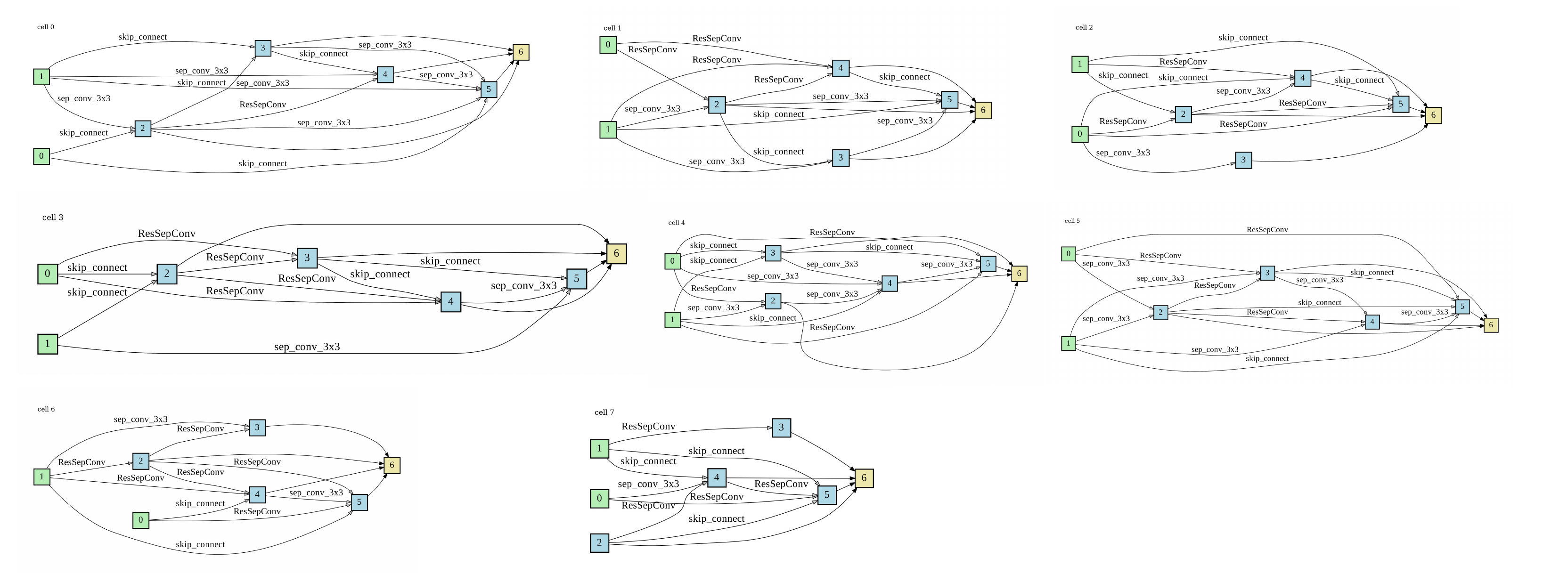}
  \caption{Architecture of MSRobNet-1560.}
  \label{fig:msrobnet-1560}
\end{figure*}

\begin{figure*}[tb]
  \centering
  \includegraphics[width=\textwidth]{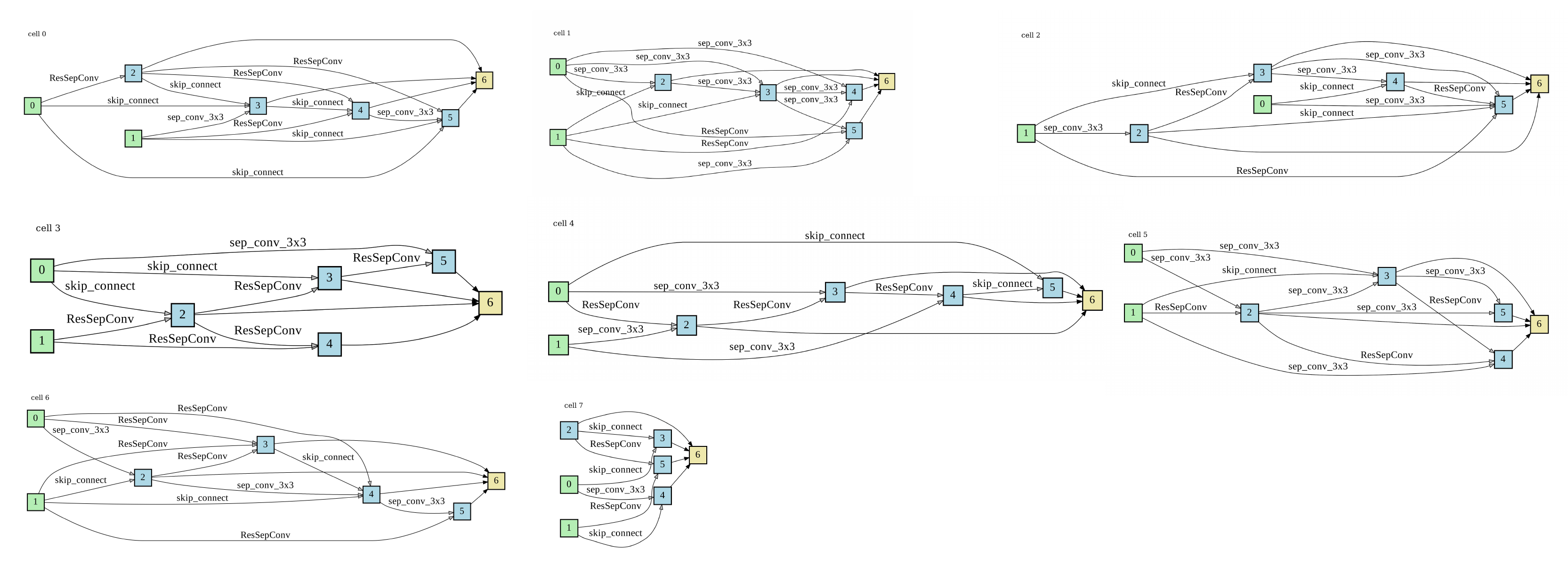}
  \caption{Architecture of MSRobNet-2000.}
  \label{fig:msrobnet-2000}
\end{figure*}

\begin{figure*}[tb]
  \centering
  \includegraphics[width=\textwidth]{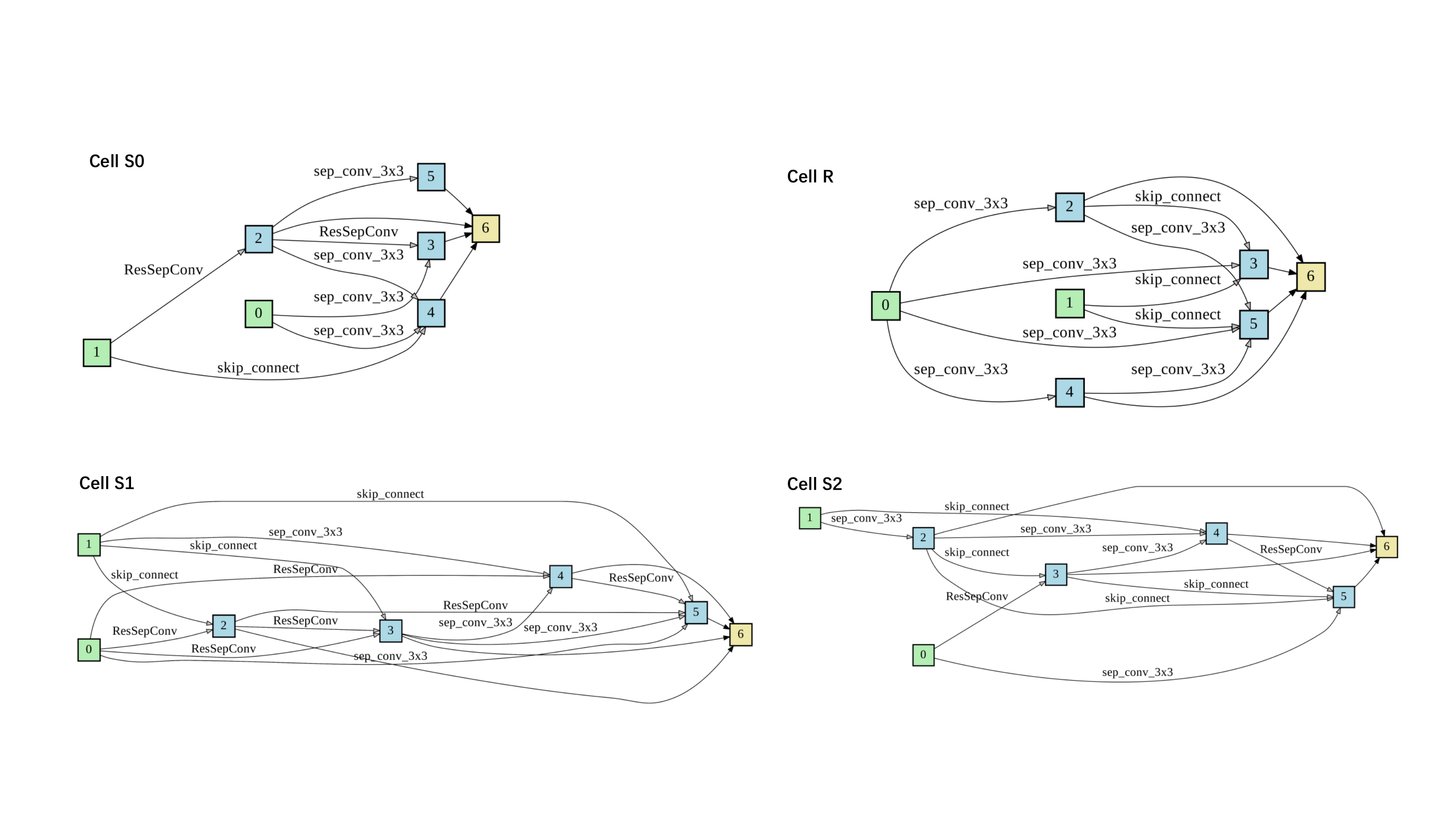}
  \caption{Architecture of MSRobNet-1560-P.}
  \label{fig:msrobnet-1560-P}
\end{figure*}

\begin{figure*}[tb]
  \centering
  \includegraphics[width=\textwidth]{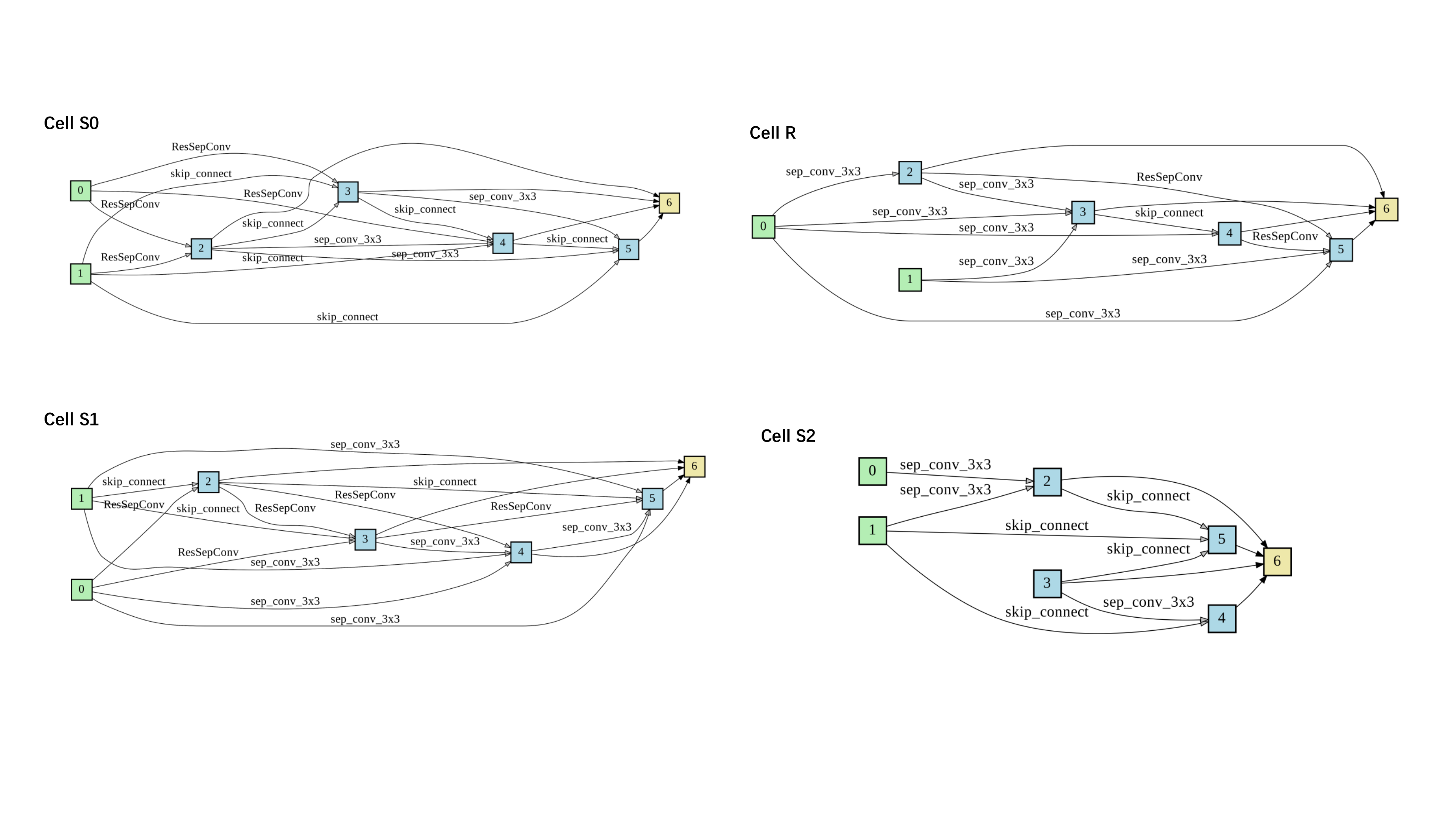}
  \caption{Architecture of MSRobNet-2000-P.}
  \label{fig:msrobnet-2000-P}
\end{figure*}

The discovered cell-wise architectures, MSRobNet-1000, MSRobNet-1560 and MSRobNet-2000, are shown in Fig.~\ref{fig:msrobnet-1000}, Fig~\ref{fig:msrobnet-1560}, and Fig.~\ref{fig:msrobnet-2000}, respectively. The discovered stage-wise architectures, MSRobNet-1560-P and MSRobNet-2000-P, are shown in Fig.~\ref{fig:msrobnet-1560-P} and Fig.~\ref{fig:msrobnet-2000-P}, respectively.

\end{appendices}
\end{document}